# Development and Validation of a Dynamic Kidney Failure Prediction Model Based on Deep Learning: A Real-World Study with External Validation

Jingying Ma[a,b,1], Jinwei Wang[c,d,e,1], Lanlan Lu[f], Zhiqin Jiang[g], Mengling Feng[b,h], Feifei Zhang[a], Peng Shen[g], Yexiang Sun[g,2], Shenda Hong[a,i,j,2], Luxia Zhang[a,c,d,e,j,2]

[a]*National Institute of Health Data Science, Peking University, 38 Xueyuan Road, Haidian District, Beijing 100191, Beijing, China*
[b]*Saw Swee Hock School of Public Health, National University of Singapore, 12 Science Drive 2, Singapore 117549, Singapore, Singapore*
[c]*Renal Division, Department of Medicine, Peking University First Hospital, No.8 Xishiku Street, Xicheng District, Beijing 100034, Beijing, China*
[d]*Research Units of Diagnosis and Treatment of Immune-Mediated Kidney Diseases, Chinese Academy of Medical Sciences, No.9 Dongdan Santiao, Dongcheng District, Beijing 100730, Beijing, China*
[e]*State Key Laboratory of Vascular Homeostasis and Remodeling, Peking University, 38 Xueyuan Road, Haidian District, Beijing 100191, Beijing, China*
[f]*Xiaying primary health care center, Ningbo Yinzhou No.2 Hospital, No.1092 Qianhe North Road, Ningbo 315192, Zhejiang, China*
[g]*Yinzhou District Center for Disease Control and Prevention, No.1221 Xueshi Road, Ningbo 315040, Zhejiang, China*
[h]*Institute of Data Science, National University of Singapore, 3 Research Link, Singapore 117602, Singapore, Singapore*
[i]*NHC Key Laboratory of Cardiovascular Molecular Biology and Regulatory Peptides, Peking University, 38 Xueyuan Road, Haidian District, Beijing 100191, Beijing, China*
[j]*Institute for Artificial Intelligence, Peking University, No. 5 Yiheyuan Road, Haidian District, Beijing 100871, Beijing, China*

[1] Equal Contributions.
[2] Corresponding Authors.
Email addresses: zhanglx@bjmu.edu.cn (Luxia Zhang), hongshenda@pku.edu.cn (Shenda Hong), 19464337@qq.com (Yexiang Sun)

## Abstract

**Background:** Chronic kidney disease (CKD), a progressive disease with high morbidity and mortality, has become a significant global public health problem. Most existing models are static and fail to capture temporal trends in disease progression, limiting their ability to inform timely interventions. We address this gap by developing a dynamic model that leverages common longitudinal clinical indicators from real-world electronic health records (EHRs) for real-time kidney failure prediction.

**Methods:** We developed KFDeep, a deep learning model built upon a time-aware long short-term memory framework that dynamically predicts kidney failure progression in individuals with CKD stages 3 to 5 by leveraging longitudinal clinical data. The model uses eight predictors, including two demographic factors (age and gender) and six repeated common clinical indicators (estimated glomerular filtration rate (eGFR), urine albumin-to-creatinine ratio (uACR), albumin levels, serum calcium, serum phosphate, and bicarbonate ($HCO_3$) levels). We evaluated its performance using the area under the receiver operating characteristic curve (AUROC) with 95% confidence intervals (CIs), alongside complementary metrics. Furthermore, we conducted subgroup analyses, interpreted the model using SHapley Additive exPlanations (SHAP) and assess whether the learned representation clusters aligned with established medical knowledge. To support clinical use, we derived explicit risk equations from the trained model for dependency-free deployment.

**Findings:** A retrospective cohort of 4,587 patients from the CK-NET-Yinzhou Dataset was used for model development (2,752 patients for training, 917 patients for validation) and internal validation (918 patients). External validation was performed in three cohorts: the prospective PKUFH cohort (934 patients), the C-STRIDE cohort (1,570 patients), and the iCaReMe cohort (498 patients). The model demonstrated competitive performance across the internal and three external validation cohorts, achieving AUROCs of 0.9311 (95% CI, 0.8873-0.9749), 0.8141 (0.7728-0.8554), 0.8427 (0.8213-0.8641), and 0.9359 (0.9031-0.9687), respectively. The model also demonstrated progressively improving dynamic predictions, good calibration, and clinically consistent interpretability. KFDeep has been deployed on an open-access website and in primary care settings.

**Interpretation:** The KFDeep model enables dynamic prediction of kidney failure without increasing clinical examination costs. It has been integrated into existing hospital systems, providing physicians with a continuously updated decision-support tool in routine care.

**Funding:** National Natural Science Foundation of China.

## Research in context

**Evidence before this study:** We conducted a literature search on Google Scholar and PubMed using keywords including '*kidney failure prediction*', '*renal failure prediction*', and '*CKD patients*', without language restrictions. We found that most studies used the static Kidney Failure Risk Equation (KFRE) or employed simple machine learning methods to model data from a single patient visit, without considering the relationships between repeated measurements over time. In addition, many studies have confirmed that estimated glomerular filtration rate (eGFR) and urine albumin-to-creatinine ratio (uACR) are the most critical indicators for predicting kidney failure.

**Added value of this study:** This study introduces a dynamic, real-time kidney failure prediction model that uses the same common clinical indicators as KFRE while incorporating longitudinal information from multiple clinical visits through advanced deep learning techniques. The model demonstrated superior performance over established prediction models across both internal and external validation cohorts. Furthermore, we provided interpretability for this model, confirming that eGFR and uACR are crucial predictors of kidney failure. Finally, we make all model parameters publicly available and have developed a web-based calculator, enabling deployment in primary healthcare settings without requiring any specific deep learning framework, and allowing direct use by clinicians.

**Implications of all the available evidence:** Our research offers a novel dynamic prediction model for assessing whether CKD patients will progress to kidney failure, with dependency-free deployment. This facilitates the direct application of AI tools in primary healthcare facilities.

## Introduction

Chronic kidney disease (CKD) is a major public health problem, with a prevalence of 10%-16% in major countries and a rising mortality trend, projected to become the fifth leading cause of death by 2040 [1,2]. Moreover, some patients will eventually progress to kidney failure, requiring lifelong dialysis or kidney transplantation to sustain life. These therapies substantially reduce quality of life and increase medical expenses [3,4]. Since CKD is asymptomatic in its early stages yet incurable once it reaches moderate to advanced stages, early diagnosis and timely intervention are of paramount importance [5].

Prediction models play a crucial role in forecasting CKD progression and guiding optimal management strategies. Among these, the Kidney Failure Risk Equations (KFREs) are the most well-validated across diverse populations worldwide [6,7]. The simple form of KFRE includes age, gender, estimated glomerular filtration rate (eGFR), and urine albumin-to-creatinine ratio (uACR), while the extended form adds four serum laboratory measures. Although AI models can capture complex interactions and collinearity among variables, existing machine learning approaches have shown limited improvement over KFRE in practice [8]. This highlights the need for deep learning models that can better capture complex patterns and learn directly from large datasets.

Moreover, research has shown that incorporating repeated measurements of indicators such as eGFR can significantly improve the accuracy of prediction models for CKD progression [9]. However, most existing models are static, relying on single time-point data and missing the benefits of tracking changes over time [6,7]. Deep learning models, particularly recurrent neural networks like long short-term memory (LSTM) networks, have a clear advantage in capturing temporal patterns, making them well suited for modeling repeated measurements in longitudinal health data. Although these models have achieved success in various medical applications [10,11], they are still rarely applied to kidney failure prediction. This underscores the need for dynamic deep learning models that can fully utilize longitudinal data to provide more accurate and clinically relevant predictions.

Therefore, we develop KFDeep, a time-aware dynamic kidney failure prediction model that incorporates repeatedly measured, commonly available clinical indicators of CKD and provides real-time risk predictions. We assess model bias through calibration curves, decision curve analysis, and subgroup analyses, and evaluate interpretability using SHAP analysis and hidden-layer clustering. Additionally, we developed a publicly accessible web-based calculator

(https://visdata.bjmu.edu.cn/kfdeep) and released the code at https://github.com/PKUDigitalHealth/KFDeep.

## Methods

This study aimed to develop a dynamic model for predicting kidney failure using the same predictors as the Kidney Failure Risk Equations (KFREs). The overall workflow is summarized in Figure 1A. We conducted the study in four steps: development, evaluation, robustness analysis, and interpretation. First, we developed the model using the CK-NET-Yinzhou cohort, an electronic health record (EHR)-based dataset from a coastal city in eastern China [12]. Second, we performed internal and external validation, assessing calibration and net benefit through decision curve analysis. Third, subgroup analyses were conducted by gender and age. Finally, we applied SHAP for model interpretability and clustered deep features, using dimensionality reduction for visualization. The study was performed and reported in accordance with the TRIPOD (Transparent Reporting of a Multivariable Prediction Model for Individual Prognosis or Diagnosis) guideline.

### Study population

The study has five source populations. For model development and internal validation, we used a retrospective cohort, the CK-NET-Yinzhou study, a real-world, electronic health record (EHR)–based cohort of permanent residents registered in the Yinzhou District Regional Health Information System in Ningbo, Zhejiang, China. The Yinzhou Health Information System integrates multiple data streams, including public health management, health examinations, outpatient visits, and hospitalizations [13]. Between May 1, 2008, and December 31, 2019, we identified 85,519 patients with CKD among 976,409 individuals [14,15]. To satisfy the chronicity requirement for CKD, we included patients with eGFR ≤60 mL/min/1.73 m² on at least two occasions separated by ≥3 months and ≤2 years. The date of the second qualifying eGFR ≤60 mL/min/1.73 m² was assigned as the index date. This index date was used only to assess cohort eligibility and to anchor baseline models based on a single set of measurements; it was not used as the time-zero point for the KFDeep longitudinal input sequence. We excluded patients who were receiving maintenance dialysis or had undergone kidney transplantation by the index date, as well as those who initiated either treatment within 3 months after the index date, given the high likelihood that they had already progressed to kidney failure before the index date. The study was approved by the Institutional Review Board of Peking University First Hospital (Approval ID:

2019[24], granted specifically for the CK-NET-Yinzhou dataset). Given the retrospective, data-only nature of the study, a consent waiver was granted.

For external validation, we used three independent CKD cohorts. The first was the Peking University First Hospital (PKUFH) CKD cohort [16], a prospective single-center cohort. Peking University First Hospital established China's first kidney division and is a national kidney-disease reference center, serving patients from across the country. From January 2013 to December 2021, 1,115 adults with CKD stages 3-5 were enrolled and followed up every 3-6 months. After applying inclusion and exclusion criteria consistent with the development cohort, 934 patients remained. Written informed consent for future research use of clinical data was obtained from all participants at enrollment. The study was approved by the Ethics Committee of Peking University First Hospital (Approval ID: 2011[363], granted specifically for the PKUFH dataset).

The second external cohort was the Chinese Cohort Study of Chronic Kidney Disease (C-STRIDE) [17], a nationwide, multicenter, prospective cohort that enrolled approximately 3,700 adults with non-dialysis CKD from 39 clinical centers across China, beginning in November 2011. 1,570 eligible participants were included for external validation after applying the study eligibility criteria. The C-STRIDE protocol was approved by the Ethics Committee of Peking University First Hospital (Approval ID: 2011[363]) and the ethics committees of the participating centers, and all participants provided written informed consent.

The third external cohort was derived from the CKD arm of iCaReMe China [18], a prospective, multicenter, observational real-world registry conducted across China. The initial CKD data extract comprised 6,000 adults. Participants were recruited between July 2023 and December 2024. The final iCaReMe external validation cohort comprised 498 participants. The registry protocol was approved by the Ethics Committee of Peking University First Hospital (Approval ID: 2024 [296-001]) and each participating center, and all participants provided written informed consent. The participant-selection processes for the CK-NET-Yinzhou development cohort and all three external validation cohorts are shown in Figure 1B.

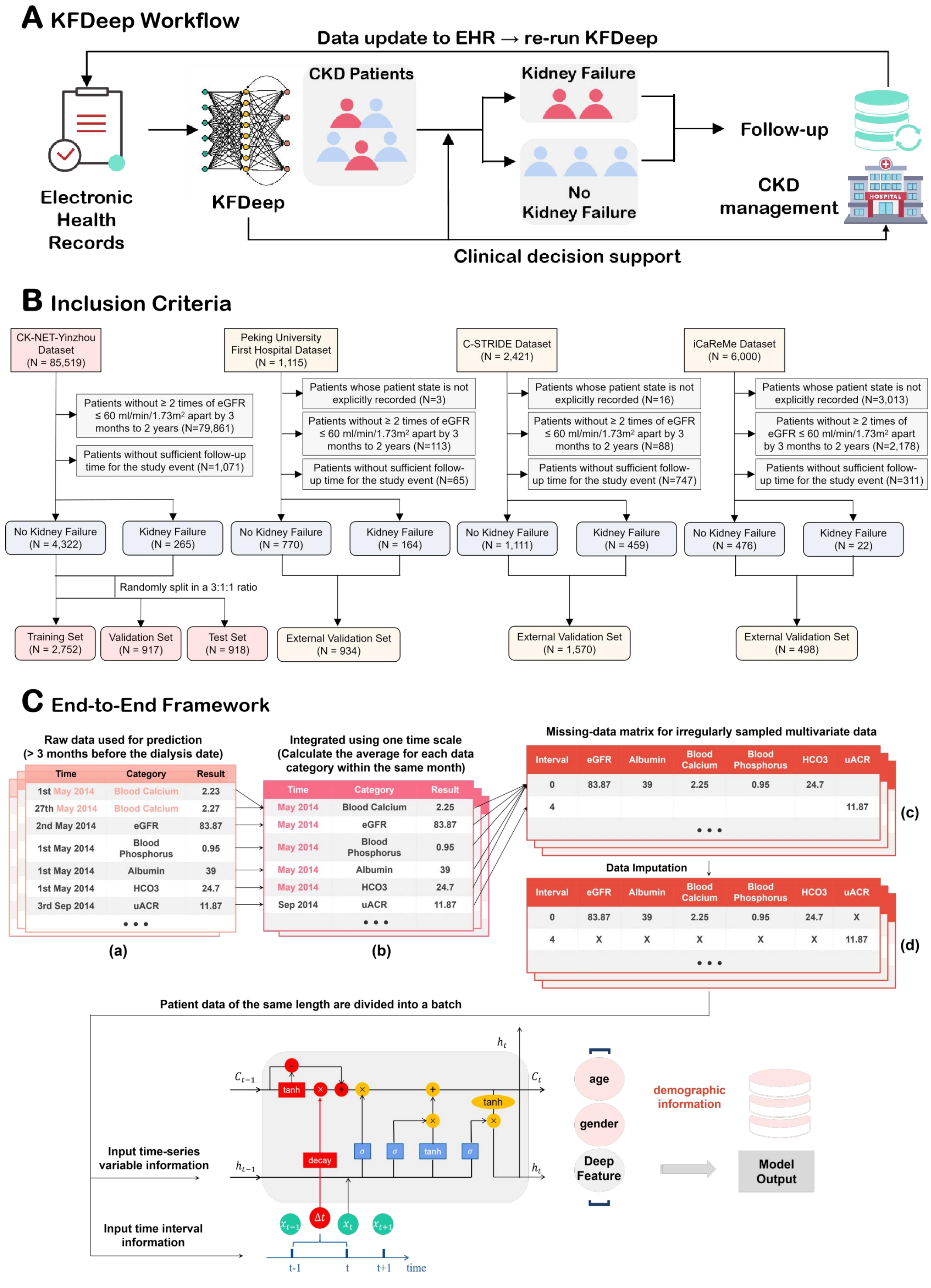

Figure 1: Overall study design of the KFDeep model. (A) KFDeep workflow. Electronic health record (EHR) data feed into KFDeep, which classifies each patient as kidney failure (KF) or non-KF. Follow-up data are appended to the EHR and trigger re-evaluation. The outputs provide clinical decision support for clinician-led CKD management. (B) Inclusion/exclusion flowchart for the internal and external cohorts. (C) End-to-end framework from data pre-processing to the KFDeep architecture, starting from each patient's earliest available measurements through to integration, data imputation, and model prediction.

In addition, we performed a single time-point validation using the UK Biobank (UKBB) [19], a large, population-based cohort with publicly accessible data. From 502,369 participants, we identified 3,261 individuals with CKD who met criteria consistent with those applied to our primary datasets. Unlike the previous cohorts, which are EHR-based with multiple time points per patient, the UKBB contains only baseline measurements, which does not allow dynamic prediction but provides an opportunity to assess model generalizability in a distinct, predominantly European population. Details of participant selection are provided in Appendix K.

## Predictors and outcomes

We select the eight variables in the KFREs for analysis [6], including six time-series variables (eGFR, urine albumin-to-creatinine ratio [uACR], albumin, serum calcium, serum phosphate, and HCO3), and two static variables (age and gender). All laboratory measurements, including serum creatinine and proteinuria, were extracted exclusively from outpatient visits, with no hospitalisation data used, to reflect routine CKD management and avoid potential bias from acute kidney injury (AKI) events. eGFR is calculated using the serum creatinine-based CKD-EPI equation [20], detailed in Appendix A. In the CK-NET-Yinzhou dataset, when direct uACR measurements were unavailable, uACR-equivalent values were derived from observed dipstick urinary protein or urine protein-to-creatinine ratio using published conversion equations [21], as detailed in Appendix B.

This study takes the initiation of kidney replacement therapy, including maintenance dialysis (hemodialysis or peritoneal dialysis) or kidney transplantation, as the outcome. For the retrospectively assembled CK-NET-Yinzhou cohort, the outcome is identified according to the service items in medical billing with hemodialysis identified by claims records of hemodialyzer and related operations and peritoneal dialysis from claims for peritoneal dialysis fluid. ICD codes are also used to ascertain dialysis-related diagnoses, as discussed in Appendix C.1-C.2. If a diagnosis of acute kidney injury was detected (related ICD-10 codes are listed in Appendix C.3), concurrent dialysis would not be labeled as an outcome. In prospectively followed PKUFH, C-STRIDE, and iCaReMe cohorts, the outcome was ascertained from prospectively recorded clinical follow-up, with the first documented initiation of dialysis or kidney transplantation used as the event date. For each participant, a single prediction was generated using information available up to the participant-specific prediction date, and the outcome was defined over the subsequent 1 year. A participant was labeled as KF if kidney replacement therapy was initiated within this period and as non-KF only when at least 1 year of follow-up confirmed the absence of kidney replacement therapy. For UKBB, kidney replacement therapy initiation was identified from hospital admission records and procedure codes, as detailed in Appendix K.

## Data pre-processing

Data pre-processing included temporally truncating the input sequences, defining the time scale, restructuring data format, and imputing missing values. First, to prevent data leakage, records within 3 months before KRT initiation or thereafter were excluded for participants with KF, whereas post-prediction records for those without KF were used only for 1-year outcome ascertainment. Second, we set the time scale to one month and averaged multiple records within each unit (Figure 1C, a→b). The rationale for this choice is discussed in Appendix E. For KFDeep, Visit 0/Interval 0 was defined as the month containing the earliest available measurement in the longitudinal input sequence. Subsequent intervals were indexed monthly from this patient-specific sequence origin. Third, data were reshaped into a multivariate, irregularly sampled matrix with missing values (b→c). Fourth, missing values were imputed using forward fill, backward fill, or interpolation based on their position (c→d), with further details provided in Appendix F.

## Model development

The CK-NET-Yinzhou dataset was randomly split into training, validation, and test sets in a 3:1:1 ratio, considering factors such as age, gender, occurrence of kidney failure, and mortality. Statistical comparisons of variable distributions (chi-square test for categorical variable and ANOVA for continuous variable) were conducted across the three datasets, revealing no significant difference. The split was performed at the patient level, ensuring that all repeated measurements from the same patient were assigned to a single subset to prevent data leakage. Then, six time series variables from each patient, along with their time intervals, were fed into the time-aware long short-term memory model (T-LSTM) model [22] in a batch-wise manner.

T-LSTM is an extension of the traditional Long Short-Term Memory (LSTM) model designed to handle irregular time intervals in time series data. In LSTM, the cell state $C_t$ is updated using the following formula:

$$C_t = f_t * C_{t-1} + i_t * \widetilde{C}_t,$$

Here, $C_{t-1}$ is the previous cell state; $f_t$ is the forget gate, which determines how much of the previous state is retained in the current state; $i_t$ is the input gate, which controls the extent to which the current input affects the cell state; and $\widetilde{C}_t$ is the candidate cell state, computed based on the current input and the previous hidden state. Unlike standard LSTM, T-LSTM introduces a time decay mechanism to adjust the memory cell states based on the elapsed time between observations. The key modification introduces a time gate that modulates the LSTM cell states, and the updated cell state $\widetilde{C}_t$ is computed as follows:

$$\widetilde{C}_t = C_t + f_t * (g(\Delta t)\text{-}1)C^S_{t\text{-}1},$$

where  represents the time decay factor,  is the time gap between consecutive observations, and  is learnt from . This mechanism allows T-LSTM to capture time-sensitive patterns in patient data more effectively. More details are explained in Appendix M-N.

The output from the T-LSTM model is a high-dimensional vector, which serves as the patient's deep feature. This deep feature is then concatenated with two static patient variables (age and gender). The concatenated vector is fed into a fully connected neural layer to get the final prediction value between 0 and 1.

## Model evaluation

To evaluate the model performance, we choose the area under the receiver operating characteristic curve (AUROC) with 95% confidence interval (CI) as the primary metric. Additional metrics included average precision (AP), sensitivity, specificity, positive predictive value (PPV), negative predictive value (NPV), F1-score, balanced accuracy, Brier score and expected calibration error (ECE).

We selected established clinical models developed specifically for kidney failure prediction as baselines, including the four- and eight-variable static KFREs [6], the latest-available-measurement Cox model (dynamic Cox) [23], and the shared-parameter joint model [25]. The prediction horizon was set to 1 year for the dynamic Cox and joint models to match that of KFDeep, whereas the established 2-year KFREs were used as the closest available fixed-horizon comparators. Detailed descriptions are provided in Appendix G. A standard LSTM was evaluated separately as an ablation of the time-aware mechanism, as described in Appendix O. To evaluate how model performance changed as longitudinal information accumulated, we compared KFDeep with the dynamic Cox and shared-parameter joint models using progressively increasing numbers of cumulative visits. This analysis was conducted in CK-NET-Yinzhou, PKUFH, and C-STRIDE, which provided sufficient longitudinal follow-up and repeated measurements for visit-stratified evaluation.

Beyond the quantitative comparisons with baseline models, we further assessed the proposed model using several graphical analyses. A calibration curve was generated to evaluate the agreement between predicted risks and observed outcomes. Additionally, to evaluate clinical utility, decision curve analysis (DCA) was conducted, comparing the net benefit of using the model versus “treat all” or “treat none” strategies across a range of threshold probabilities.

To examine the robustness of the proposed model, we additionally evaluated its subgroup performance on the internal test set based on age (18-59, 60-74, ≥75 years) and gender (male, female), and reported detailed results from five-fold cross-validation.

### Model interpretation

We employed the SHapley Additive exPlanations (SHAP) method to calculate feature importance in the proposed model and to determine if the correlation between input features and the output is positive or negative. Moreover, we delved into the practical implication of the deep features. We used the k-means algorithm to cluster the hidden layer vectors and applied the T-SNE method to reduce them to a 2-dimensional space and visualize the clustering results. In addition, we utilized radar and box plots to analyze the differences in variables and model predictions among the categories obtained from clustering.

### Model deployment

We explicitly extract the model weights and transform them into a set of visible equations (see Appendix R), making the model fully interpretable and independent of any specific deep learning architecture. KFDeep has been deployed on a web-based platform, allowing users to input a patient’s historical data for all eight indicators and obtain the predicted risk of kidney failure progression. In addition, the model has also been implemented in primary care settings to assist clinicians in decision-making.

### Role of the funding source

The funders of the study had no role in the study design, data collection, data analysis, data interpretation, or writing of the report.

## Results

A total of 4,587 patients from CK-NET-Yinzhou were included for model development and internal validation, comprising 2,752 patients in the training set, 917 in the validation set, and 918 in the test set. External validation included 934 patients from PKUFH, 1,570 from C-STRIDE, and 498 from iCaReMe. Table 1 summarizes the overall baseline characteristics of the four study cohorts; detailed cohort-specific characteristics, comparisons across the stratified internal data splits, and predictor availability are provided in Appendix D.

Table 1: Baseline Characteristics of the Study Cohorts.

| | **Development set (CK-NET-Yinzhou) (N = 4587)** | **External validation (PKUFH) (N = 934)** | **External validation (C-STRIDE) (N = 1570)** | **External validation (iCaReMe) (N = 498)** |
|---|---|---|---|---|
| **Demographics** | | | | |
| Age (years) | 77 (68,83) | 68 (57, 81) | 54 (43, 64) | 64 (55, 71) |
| Male, n (%) | 2398 (52) | 493 (53) | 886 (56) | 303 (61) |
| History of Diabetes, n (%) | 1516 (33) | 184 (20) | 308 (20) | 175 (35) |
| Body Weight (kg) | 62.5 (56.1, 70.0) | 64.8 (59.1, 72.5) | 65.0 (57.0, 74.0) | 65.0 (58.0, 75.0) |
| Systolic BP (mmHg) | 152 (148, 160) | 154 (136, 170) | 131 (121, 142) | 132 (121, 144) |
| Diastolic BP (mmHg) | 90 (88, 98) | 92 (76, 107) | 80 (75, 88) | 80 (70, 86) |
| **Clinical and Laboratory Characteristics** | | | | |
| eGFR (mL/min/1.73m$^2$) | 55 (44, 59) | 33 (20, 47) | 32 (23, 43) | 38 (24, 49) |
| uACR (mg/g) | 11.9 (11.9, 25.2) | 197.8 (25.2, 579.5) | 369.5 (69.4, 946.4) | 264.7 (70.2, 748.5) |
| Albumin (g/L) | 40.1 (36.4, 43.7) | 42.8 (39.9, 45.1) | 40.9 (36.7, 44.0) | 41.7 (38.1, 44.2) |
| Serum Calcium (mmol/L) | 2.2 (2.1, 2.3) | 2.3 (2.2, 2.4) | 2.2 (2.1, 2.4) | 2.3 (2.2, 2.4) |
| Serum Phosphorus (mmol/L) | 1.02 (0.91, 1.15) | 1.19 (1.06, 1.35) | 1.20 (1.06, 1.36) | 1.21 (1.06, 1.38) |
| $HCO_3$ (mmol/L) | 24.1 (21.4, 27.0) | 24.4 (22.2, 26.4) | 24.7 (22.0, 27.5) | 24.3 (21.6, 26.3) |
| Fasting Plasma Glucose (mmol/L) | 6.14 (5.33, 7.17) | 5.58 (4.99, 6.23) | 5.08 (4.54, 5.79) | 5.59 (5.00, 6.58) |
| Total Cholesterol (mmol/L) | 4.57 (3.79, 5.40) | 4.56 (3.83, 5.44) | 4.64 (3.89, 5.43) | 4.55 (3.91, 5.53) |
| LDL-C (mmol/L) | 2.55 (1.98, 3.16) | 2.41 (1.77, 2.93) | 2.57 (2.06, 3.15) | 2.59 (2.04, 3.24) |
| HDL-C (mmol/L) | 1.13 (0.95, 1.36) | 1.19 (1.00, 1.41) | 1.06 (0.90, 1.32) | 1.08 (0.90, 1.29) |
| Triglycerides (mmol/L) | 1.35 (0.94, 1.94) | 1.24 (0.87, 1.78) | 1.77 (1.21, 2.44) | 1.64 (1.17, 2.35) |
| **Outcomes** | | | | |
| Kidney Failure Events, n (%) | 266 (6) | 164 (17) | 459 (29) | 22 (4) |
| Dialysis, n (%) | 227 (85) | 162 (98) | 440 (96) | 22 (100) |
| Transplantation, n (%) | 39 (15) | 2 (2) | 19 (4) | 0 (0) |
| Mortality, n (%) | 777 (17) | 14 (1.5) | 79 (5) | NA |
| **Follow-up Information** | | | | |
| Visits (times) | 8 (5,13) | 11 (5, 23) | 5 (3, 7) | 3 (2, 3) |
| Observation time (y) | 5.57 (3.58, 7.41) | 3.33 (1.75, 5.75) | 2.43 (1.61, 3.70) | 1.00 (1.00, 1.00) |

Continuous variables are presented as the median (25th and 75th interquartile range). Categorical variables are presented as counts (n) and percentages (%).

The experimental results for KFDeep and baseline models are shown in Table 2 (the KFRE baseline was not applicable to the iCaReMe dataset because its follow-up duration did not support a 2-year prediction horizon), with KFDeep achieving the highest AUROC in internal and external validations. Statistically significant improvements are marked with an asterisk. In addition to AUROC, we evaluated performance using AP, sensitivity, specificity, PPV, NPV, F1-score, balanced accuracy, Brier score, and ECE, where KFDeep consistently achieved competitive performance. Temporal comparative experiments in the three cohorts with sufficient longitudinal follow-up (Figure 2) further demonstrated that KFDeep generally maintained higher AUROC than dynamic baseline models across visit counts, with trend tests (Spearman's ρ and p-values) indicating statistically significant performance gains as more visits were incorporated. While the main analysis focused on the longitudinal datasets, we also performed single-timepoint external validation on the UKBB cohort (European) to assess generalizability under different population structures. Detailed results, including additional evaluation metrics, pairwise DeLong tests, temporal comparative experiments, and single-timepoint UKBB validation, are provided in Appendices H-K.

Table 2: AUROC Scores of Baseline Models.

| **Model** | **Internal Validation (CK-NET-Yinzhou)** | **External Validation (PKUFH)** | **External Validation (C-STRIDE)** | **External Validation (iCaReMe)** |
|---|---|---|---|---|
| ***Static models*** | | | | |
| KFRE (4 variables, 2 years) [6] | 84.50 ± 6.20 | 70.06 ± 6.05 | 76.06 ± 3.59 | - |
| KFRE (8 variables, 2 years) [6] | 86.29 ± 5.19 | 73.21 ± 5.13 | 78.92 ± 3.07 | - |
| ***Dynamic models*** | | | | |
| Latest-measurement Cox model [23] | 88.00 ± 5.10 | 74.82 ± 4.52 | <u>80.16</u> ± 2.28 | 90.26 ± 3.21 |
| Shared-parameter joint model [25] | <u>88.45</u> ± 5.23 | <u>76.53</u> ± 3.36 | 78.62 ± 2.54 | <u>91.13</u> ± 2.66 |
| KFDeep (Proposed model)* | **93.11** ± 4.38 | **81.41** ± 4.13 | **84.27** ± 2.18 | **93.59** ± 3.35 |

*Indicates statistically significant difference in AUROC compared with baseline models (DeLong test, $P < 0.05$).
The best-performing result is shown in bold, and the second-best result is underlined.

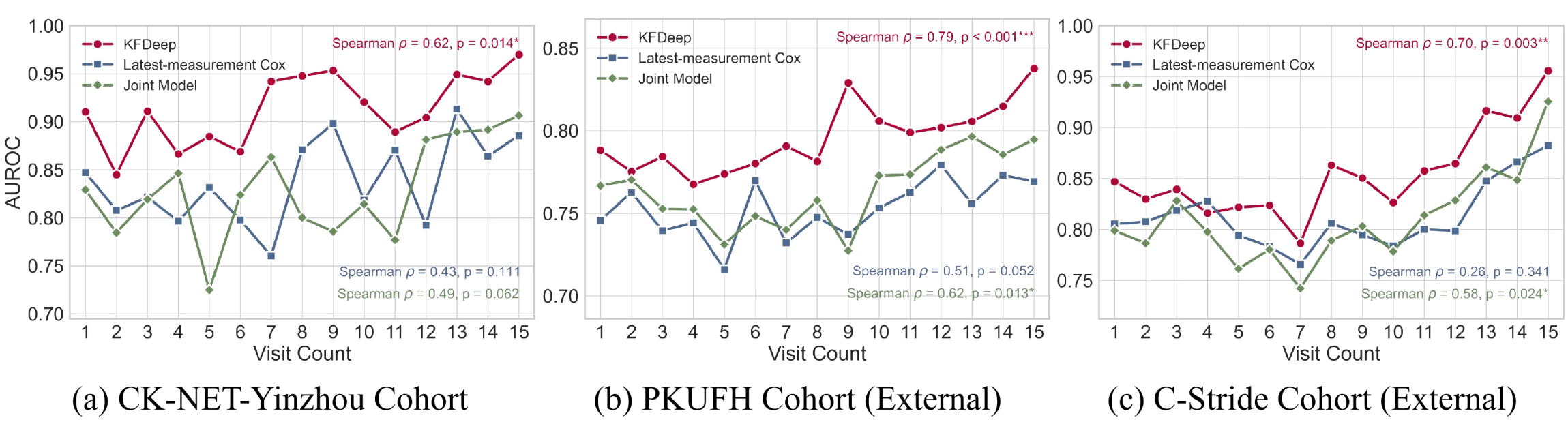

(a) CK-NET-Yinzhou Cohort (b) PKUFH Cohort (External) (c) C-Stride Cohort (External)

Figure 2: AUROC comparison of KFDeep and dynamic baseline models for predicting kidney failure within 1 year across successive prediction visits in CK-NET-Yinzhou, PKUFH, and C-STRIDE.

We further evaluated KFDeep using graphical analyses in the four longitudinal cohorts (Figure 3). Calibration generally indicates reasonable agreement between predicted outcome and observed risks, although uncertainty increased in sparsely populated intervals. Decision-curve analysis showed that KFDeep provided positive net benefit relative to the treat-none strategy and generally outperformed the treat-all strategy across a range of clinically relevant threshold probabilities. For readability, the main figure presents grouped calibration intervals together with the corresponding risk distributions; the complete 10-bin calibration plots are provided in Appendix H.

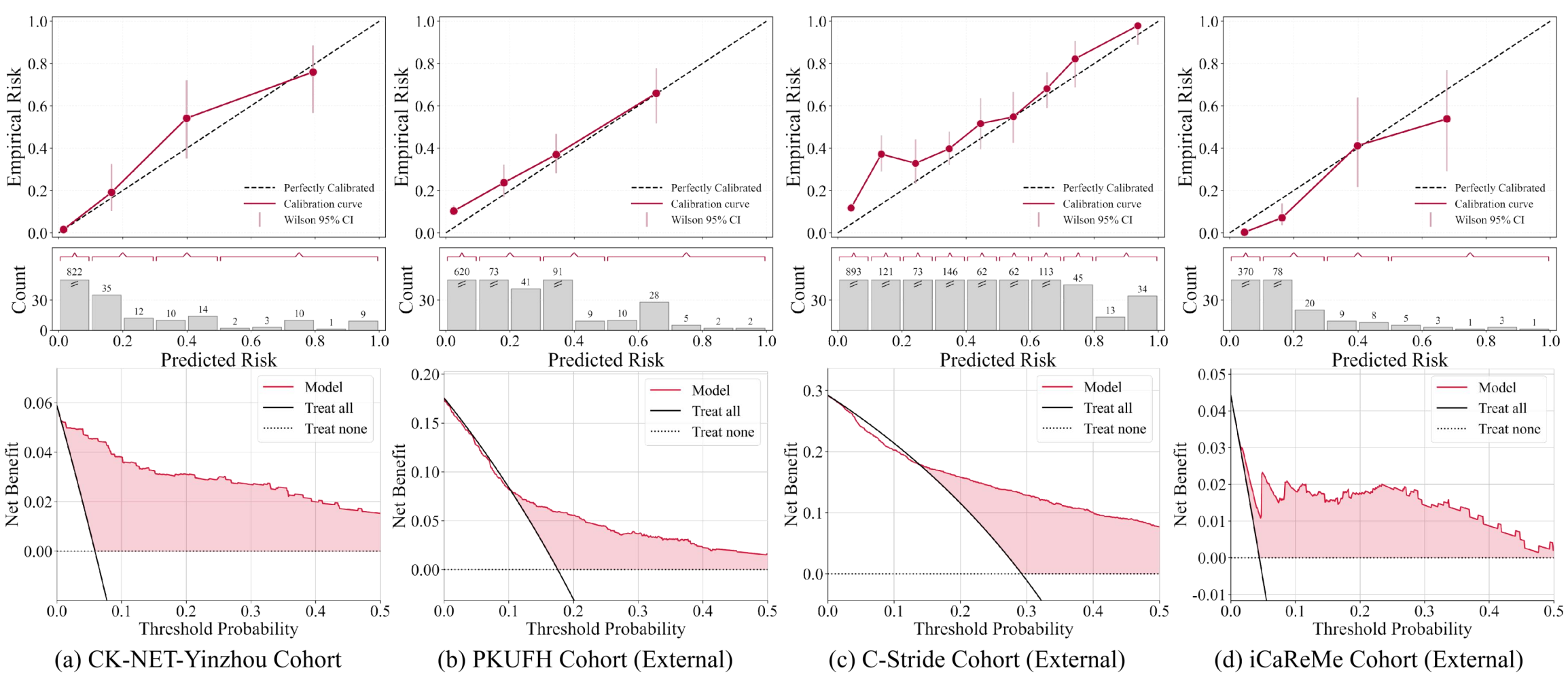

(a) CK-NET-Yinzhou Cohort (b) PKUFH Cohort (External) (c) C-Stride Cohort (External) (d) iCaReMe Cohort (External)

Figure 3: Calibration and Decision Curve Analysis Across the Internal and External Cohorts.

We assessed the robustness of KFDeep through multiple analyses. Five-fold cross-validation results demonstrated consistent performance across folds (Appendix L), and subgroup analysis by gender and age group (Figure 4) showed no statistically significant differences based on pairwise DeLong tests.

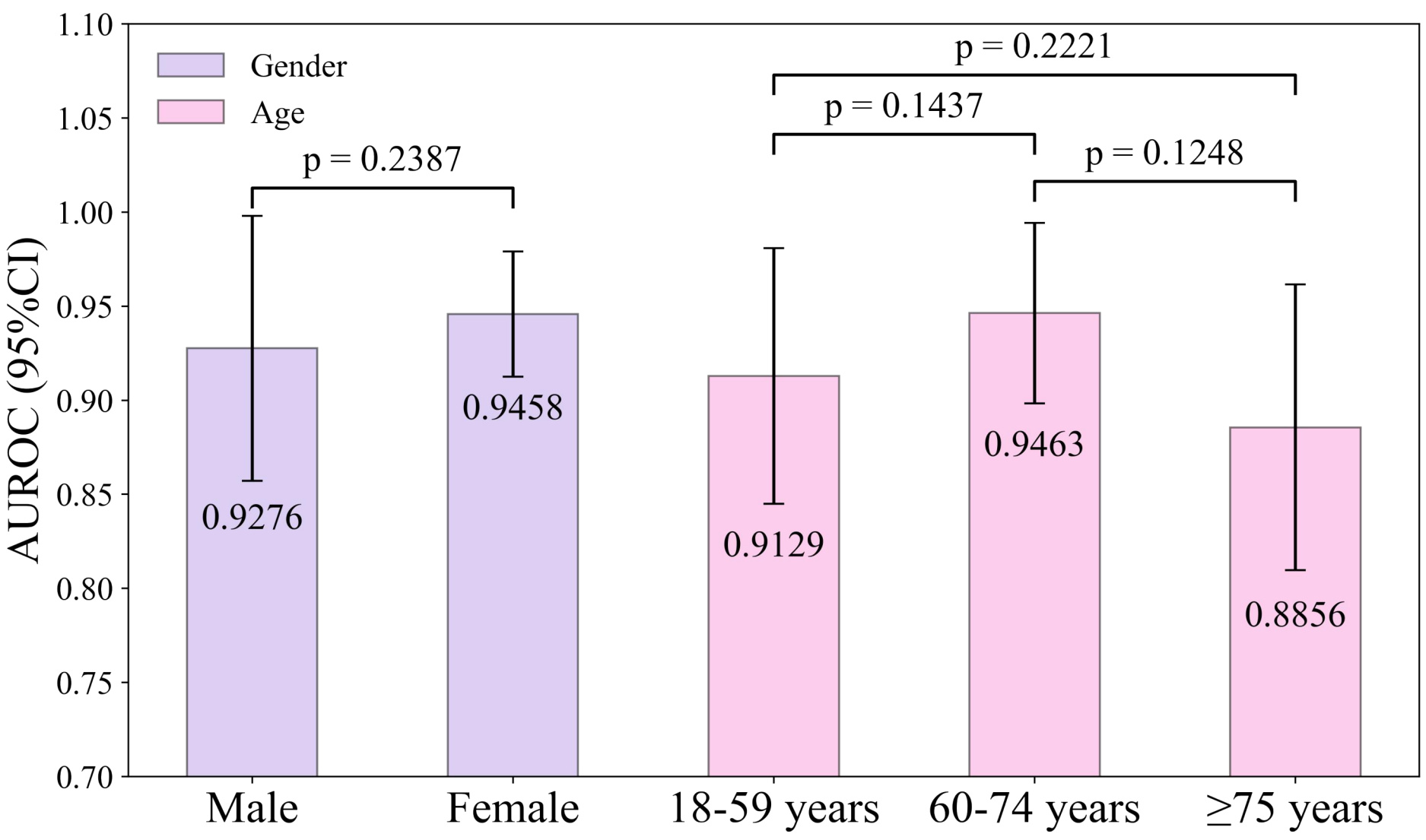

Figure 4: Subgroup Analysis.

Model interpretability was analyzed using SHAP (Figure 5), which identified uACR and eGFR as the most important predictors. To further explore representation patterns, we performed clustering on the deep features from KFDeep. The t-SNE visualization (Figure 6(a)) revealed three clusters. Cluster membership was modestly associated with CKD stage (stages 3, 4, and 5; Cramér's V=0.223). The proportion of participants with CKD stages 4-5 increased from 2.1% in cluster 1 to 7.8% in cluster 2 and 25.5% in cluster 3. Consistent with the increasing predicted risk across clusters 1-3, uACR and serum phosphorus increased, whereas eGFR decreased (Figure 6(b)). Together, these clusters suggested a graded pattern of predicted risk and clinical disease severity.

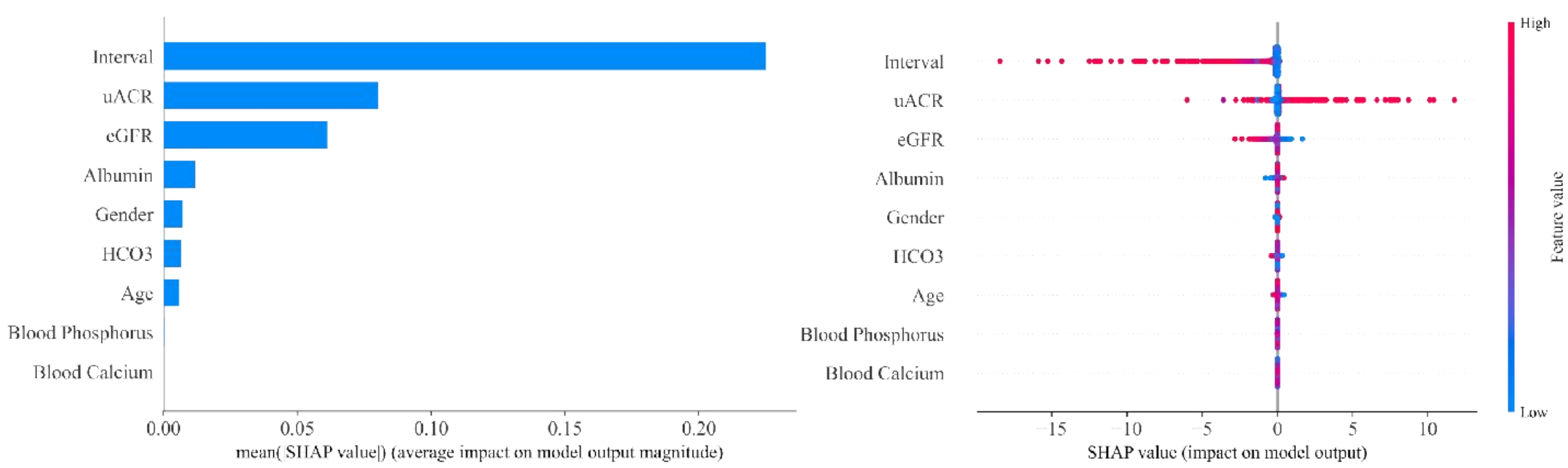

Figure 5: SHAP Analysis of Feature Importance.

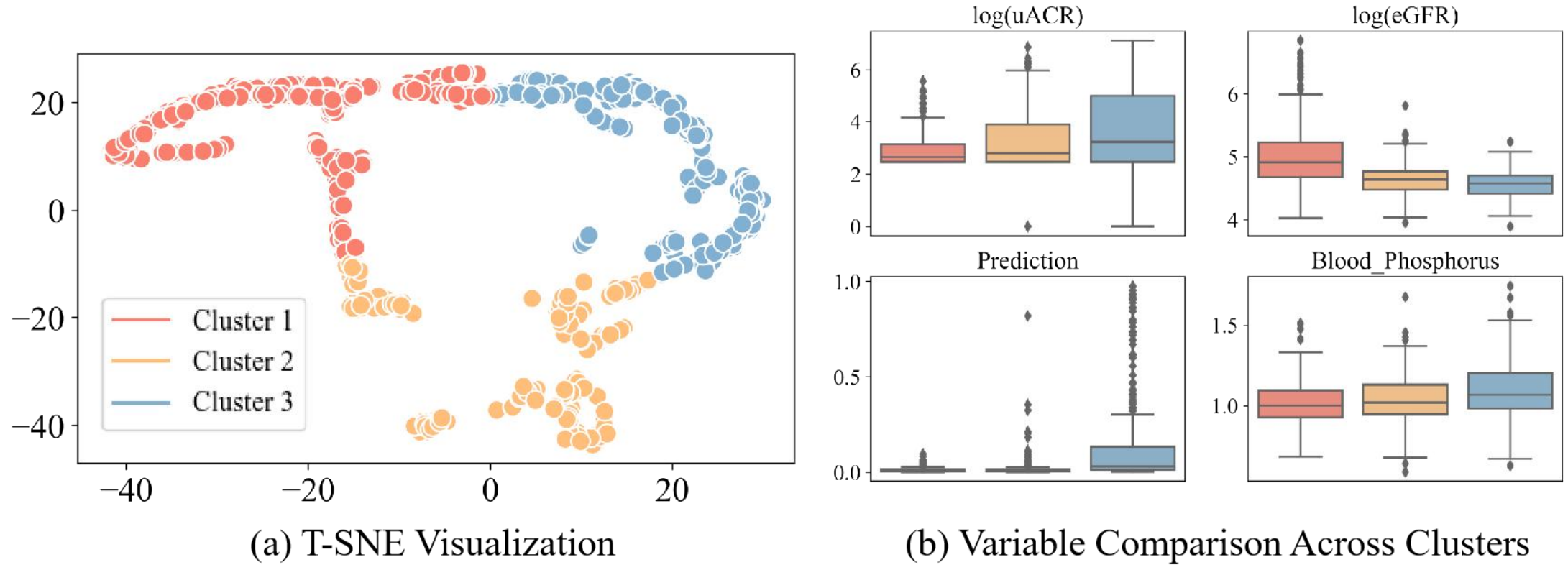

(a) T-SNE Visualization (b) Variable Comparison Across Clusters

Figure 6: Deep Features Analysis.

Beyond performance evaluation, we demonstrated the real-world applicability of KFDeep. By transforming the model into explicit formulas, we enabled deployment without reliance on deep learning frameworks. A web-based calculator was developed for direct use (available at https://visdata.bjmu.edu.cn/kfdeep). KFDeep has also been successfully implemented in primary care settings, with screenshots of the clinical interface provided in Appendix R.

## Discussion

We developed a dynamic kidney failure prediction model, KFDeep, for patients with CKD stages 3-5. The model was developed based on an EHR-based CKD registry in a coastal city of China, and externally validated in three longitudinal Chinese cohorts and a European cohort. The model retained the simplicity of the eight KFRE variables yet achieved higher accuracy than established static KFREs and dynamic prediction approaches. We have converted the model into explicit formulas and developed a web-based calculator and deployed it in primary healthcare settings, including both tertiary hospitals and community health centers, to support CKD management.

Using the same commonly available clinical indicators [24], KFDeep leverages temporal information to deliver superior performance, achieving AUROCs of 0.931 in the internal test set and 0.814, 0.843, and 0.936 in the PKUFH, C-STRIDE, and iCaReMe cohorts, respectively. Compared with the strongest baseline model in each cohort, KFDeep improved AUROC by 5.3%, 6.4%, 5.1%, and 2.7%, respectively. These improvements were statistically significant in the internal, PKUFH, and C-STRIDE cohorts. Performance remained stronger with longer follow-up windows; across all dataset, the trend test under dynamic prediction was statistically significant, indicating that performance improved as the number of visits accumulated (e.g., AUROC at 15 visits was 6.5%, 6.3%, and 12.8% higher than that at one visit in CK-NET-Yinzhou, PKUFH, and C-STRIDE, respectively), supporting the value of accumulating longitudinal information despite some visit-level fluctuations. KFDeep also achieved the highest average precision across the four longitudinal cohorts and showed generally favorable calibration and class-imbalance-sensitive performance.

External validation indicated heterogeneous transportability across three longitudinal cohorts encompassing a tertiary hospital, a nationwide prospective cohort, and a multicenter real-world registry. Compared with internal validation, AUROC decreased in PKUFH and C-STRIDE but remained comparable in iCaReMe. Kidney failure prevalence differed substantially across PKUFH, C-STRIDE, and iCaReMe (17%, 29%, and 4%, respectively), together with marked variation in baseline eGFR and uACR. These differences in case mix, rather than prevalence alone, may partly account for the observed variation in AUROC, as lower-prevalence cohorts may contain fewer but more phenotypically distinct cases. Despite the decline in two cohorts, KFDeep maintained AUROCs above 0.80 and achieved relative improvements of 2.7%-6.4% over the strongest baseline model in all three external cohorts. Its performance in the single-timepoint UKBB analysis provides complementary evidence of its potential applicability across distinct populations.

To improve transparency for potential clinical use [26,27], we examined KFDeep's feature contributions and learned representations. SHAP analysis highlights eGFR and uACR as the most influential predictors of kidney failure [28,29]. Clustering of the learned representations showed progressive enrichment of patients with CKD stages 4-5 from cluster 1 to cluster 3 (2.1% to 25.5%), accompanied by decreasing eGFR and increasing uACR, serum phosphate, and prediction [30]. These findings suggest that the representations captured clinically meaningful gradients of disease severity, with the resulting clusters not simply recapitulating conventional CKD stages; however, further validation is needed to determine whether they represent phenotypes.

Furthermore, KFDeep has a mathematically transparent architecture and can accommodate irregular intervals between visits. Based on this structure, we have developed a web-based calculator and deployed it in primary healthcare systems, including both tertiary hospitals and community health centers (Appendix R). With each new set of test results, KFDeep updates the patient's risk estimate and presents it for clinician review, providing current information to support timely, evidence-informed CKD management.

This study has several limitations. Although KFDeep was externally validated in three independent longitudinal cohorts in China, further validation in more geographically and ethnically diverse populations is warranted. Access to suitable longitudinal datasets remains limited, and broader international collaborations will be needed to support such evaluation. The lower AUROC observed in some external cohorts relative to the internal test set may reflect differences in case mix, clinical practice, and measurement patterns, as well as possible overfitting to the development population. Current deployment is limited to the Yinzhou population; for substantially different settings, local adaptation or recalibration should be considered before implementation. Death before kidney failure was not formally accounted for as a competing event, and appropriately specified competing-risk models should be evaluated in future work. Moreover, although the interpretability analyses were consistent with established clinical knowledge, they were observational and should not be interpreted causally. Future studies could distinguish predictive associations from causal relationships, evaluate the clinical impact of KFDeep through large-scale pragmatic randomized trials, and assess its cost-effectiveness and implications for health policy.

Overall, KFDeep represents a promising step toward dynamic kidney failure prediction, and its broader deployment supported by rigorous clinical assessment has the potential to improve patient outcomes.

## Conclusion

In this study, we developed and validated a temporal deep learning model that leverages repeated measurements of common clinical indicators to provide dynamic, individualized risk predictions for kidney failure in patients with CKD. Its robustness was demonstrated through external validation, along with multiple evaluation experiments. Beyond research evaluation, the model has already been deployed, including a web-based calculator and deployment in primary care settings, to support physicians in making accurate and timely patient management decisions.

### Data Availability

For the internal CK-NET-Yinzhou study dataset, the dataset cannot be made publicly available due to the terms of the agreement with the Health Commission. The external Peking University First Hospital cohort (PKUFH) dataset, Chinese Cohort Study of Chronic Kidney Disease (C-STRIDE) dataset, and iCaReMe China dataset are available upon research collaboration request to the corresponding authors. The UK Biobank (UKBB) dataset is publicly available to approved researchers through an application process via the UK Biobank Access Management System (https://www.ukbiobank.ac.uk/enable-your-research). The code and the model parameters are publicly available at https://github.com/PKUDigitalHealth/KFDeep. Additionally, we have deployed a calculator on https://visdata.bjmu.edu.cn/kfdeep for direct use.

### Author Contributors

JM, JW, and SH contributed to the study design and methodology. JW and FZ were responsible for data extraction and initial data cleaning, while JM provided data cleaning and visualization. As the CK-NET-Yinzhou study dataset (development dataset) was held within a secure environment, JM, JW, YS, PS, ZJ, SH, and LZ had access to it and were able to directly access and verify the dataset. All authors had access to the external cohorts and participated in directly accessing and verifying the data. All authors were responsible for writing, reviewing, and editing the manuscript. JM, JW, and SH contributed to the first draft. JW, LL, and LZ provided medical writing guidance, while JM, YS, MF, PS, ZJ, and SH continued to revise and enhance other parts of the manuscript. JW, SH, and LZ provided funding support. JM, JW, YS, SH, and LZ were responsible for the decision to submit the manuscript for publication.

## Declaration of Interests

All authors declare no conflicts of interest related to this research.

## Acknowledgment

SH is supported by the National Natural Science Foundation of China (62102008), Beijing Nova Program (202604841257), Beijing Municipal Science and Technology Commission (Z251100000725008). JW is supported by National High Level Hospital Clinical Research Funding (Interdepartmental Research Project of Peking University First Hospital: 2024IR14), CAMS Innovation Fund for Medical Sciences (2019-I2M-5-046) and State Key Laboratory of Vascular Homeostasis and Remodeling, Peking University. LZ is supported by the National Natural Science Foundation of China (72125009), CAMS Innovation Fund for Medical Sciences (2019-I2M-5-046) and State Key Laboratory of Vascular Homeostasis and Remodeling, Peking University. This research has been conducted using the UK Biobank Resource under application number 90018.

# Supplementary material

## Appendix A. CKD-EPI Equation

For females:

$$\text{eGFR}=\begin{cases}144\times\left(\dfrac{\text{serum creatinine}}{88.4\times0.7}\right)^{-0.329}\times0.993^{\text{age}}, & \text{if } \dfrac{\text{serum creatinine}}{88.4}\leq0.7,\\ 144\times\left(\dfrac{\text{serum creatinine}}{88.4\times0.7}\right)^{-1.209}\times0.993^{\text{age}}, & \text{otherwise.}\end{cases}$$

For males:

$$\text{eGFR}=\begin{cases}141\times\left(\dfrac{\text{serum creatinine}}{88.4\times0.9}\right)^{-0.411}\times0.993^{\text{age}}, & \text{if } \dfrac{\text{serum creatinine}}{88.4}\leq0.9,\\ 141\times\left(\dfrac{\text{serum creatinine}}{88.4\times0.9}\right)^{-1.209}\times0.993^{\text{age}}, & \text{otherwise.}\end{cases}$$

## Appendix B. Equation for uACR Conversion

The equation for calculating uACR based on dipstick urinary protein is:

$$\begin{aligned}\text{uACR}=\exp(\ & 2.4738\\ & +0.7539\times(\textbf{if}\quad \textbf{trace})\\ & +1.7243\times(\textbf{if}\quad \textbf{+})\\ & +3.3475\times(\textbf{if}\quad \textbf{++})\\ & +4.6399\times(\textbf{if}\quad \textbf{>++})).\end{aligned}$$

The equation for calculating uACR based on urine protein-to-creatinine ratio (PCR) is:

$$\begin{aligned}\text{uACR}=\exp(\ & 5.3920\\ & +0.3072 * \log(\min(\frac{PCR}{50},1))\\ & +1.5793 * \log(\max(\min(\frac{PCR}{500},1),0.1))\\ & +1.1266 * \log(\max(\frac{PCR}{500},1))).\end{aligned}$$

## Appendix C. ICD-10 Codes for Identifying Kidney Replacement Therapy Patients

### Appendix C.1. Kidney Transplant Recipients

The ICD-10 codes used for identifying kidney transplant recipients (as a potential outcome of end-stage kidney disease) are shown in Table C.1. We only include codes that indicate the completed status of kidney transplantation, excluding complications or rejection events.

Table C.1: ICD-10 codes for kidney transplant status

| Category | ICD-10 codes | |
|---|---|---|
| | **National version** | **Beijing version** |
| Kidney transplant status | Z94.000, Z94.001, Z94.002 | Z94.000, Z94.002 |
| Kidney transplant procedure | 55.611, 55.691 | 55.611, 55.691 |
| Abbreviations: ICD-10 (International Classification of Diseases-10). | | |

## Appendix C.2. Dialysis Patients

The ICD-10 codes used for identifying dialysis patients are shown in Table C.2

Table C.2: ICD-10 codes for dialysis modality

| Dialysis modality | ICD-10 codes | |
|---|---|---|
| | **National version** | **Beijing version** |
| HD | T80.801, T80.902, T82.400, T82.401, Z99.201 | T82.401, Z99.201 |
| PD | T85.609, T85.610, T85.611, T85.710, T85.711, T85.801, T85.901, Z49.201 | T80.201, T85.602, Z49.201, Z99.202 |
| Abbreviations: HD (hemodialysis), PD (peritoneal dialysis), ICD-10 (International Classification of Diseases-10). | | |

## Appendix C.3. Acute Kidney Injury

The ICD-10 codes used for identifying patients with acute kidney injury are shown in Table C.3.

Table C.3: ICD-10 Codes for Acute Kidney Injury

| ICD-10 codes | |
|---|---|
| **National version** | **Beijing version** |
| N17, N01, T79.5, D59.3, K76.7, O90.4, N96.x00, A23.100, O00.101, O00.105, O00.108, O00.111, O00.114, O02.001, O02.100, O03, O04, O05, O06.900, O07, O08 (exclude O08.006, O08.103, O08.104, O08.105, O08.106, O08.302, O08.806), | N17, N01, T79.5, D59.3, K76.7, O90.4, N96xx01, O00.102, O00.107, O00.110, O00.111, O02.101, O03, O04 (exclude O04.905, O04.907), O05, O06 (exclude O06.907, O06.908), O07, O08 (exclude |

| O20.000, O26.200, O31.100, Z09.802, Z35.100, Z35.101, Z35.104, N10.x00, N10.x01, N99.000, N99.001 | O08.103, O08.105, O08.301, O08.801, O08.803, O08.806, O08.807, O08.809), O20.001, O26.201, Z30.201, Z35.102, Z98.8308, Z98.8312, N00.908, N10xx03, N10xx04+H20.9 |
|---|---|
| Abbreviations: ICD-10 (International Classification of Diseases-10). | |

## Appendix D. Cohort-specific Baseline Characteristics

### Appendix D.1. Internal Cohort Characteristics by Data Split

The CK-NET-Yinzhou development cohort was divided into training, validation, and test sets using a stratified partitioning strategy designed to preserve comparable demographic, clinical, outcome, and follow-up distributions. The balance of the resulting partition was assessed by comparing these characteristics across the three subsets (Table D.1). No statistically significant differences were observed for any evaluated variable (all $P > 0.05$), supporting a well-balanced partition without detectable bias in the measured characteristics.

Table D.1. Baseline Characteristics of the CK-NET-Yinzhou Development Cohort by Training, Validation, and Testing Sets.

| | **Training Set** | **Validation Set** | **Testing Set** | ***P* Value** |
|---|---|---|---|---|
| **Demographics** | | | | |
| Sample Size | 2752 | 917 | 918 | - |
| Age (years) | 78 (68, 83) | 78 (69, 83) | 78 (68, 82) | 0.208 |
| Male, n (%) | 1437 (52) | 478 (52) | 483 (53) | 0.973 |
| History of Diabetes, n (%) | 908 (33) | 303 (33) | 305 (33) | 0.994 |
| Body Weight (kg) | 64.2 (56.3, 72.1) | 63.5 (55.6, 71.7) | 64.0 (56.0, 72.0) | 0.458 |
| Systolic BP (mmHg) | 156 (139, 173) | 155 (138, 172) | 157 (140, 174) | 0.392 |
| Diastolic BP (mmHg) | 91 (76, 107) | 93 (77, 108) | 92 (76, 108) | 0.636 |
| **Clinical and Laboratory Characteristics** | | | | |
| eGFR (mL/min/1.73m2 ) | 55 (44, 59) | 55 (44, 59) | 55 (44, 59) | 0.584 |
| uACR (mg/g) | 11.9 (11.9, 25.2) | 11.9 (11.9, 25.2) | 11.9 (11.9, 25.2) | 0.828 |
| Albumin (g/L) | 40.1 (36.2, 43.7) | 40.1 (36.6, 43.8) | 40.1 (36.6, 43.8) | 0.809 |
| Serum Calcium (mmol/L) | 2.2 (2.1, 2.3) | 2.2 (2.1, 2.3) | 2.2 (2.1, 2.3) | 0.871 |
| Serum Phosphate (mmol/L) | 1.02 (0.91, 1.15) | 1.02 (0.92, 1.16) | 1.02 (0.92, 1.16) | 0.423 |
| HCO3 (mmol/L) | 24.7 (24.0, 24.7) | 24.7 (24.0, 24.7) | 24.7 (24.0, 24.7) | 0.158 |
| Fasting Plasma Glucose (mmol/L) | 6.20 (5.32, 7.33) | 6.00 (5.23, 6.90) | 6.12 (5.47, 6.98) | 0.521 |
| Total Cholesterol (mmol/L) | 4.59 (3.79, 5.40) | 4.42 (3.69, 5.31) | 4.66 (3.90, 5.47) | 0.259 |
| LDL-C (mmol/L) | 2.57 (1.98, 3.18) | 2.47 (1.92, 3.08) | 2.59 (2.04, 3.19) | 0.222 |
| HDL-C (mmol/L) | 1.13 (0.95, 1.35) | 1.14 (0.95, 1.36) | 1.13 (0.95, 1.39) | 0.744 |
| Triglycerides (mmol/L) | 1.35 (0.94, 1.96) | 1.32 (0.92, 1.86) | 1.36 (0.96, 1.94) | 0.434 |

| **Outcomes** | | | | |
|---|---|---|---|---|
| Kidney Failure events, n (%) | 159 (5.8) | 53 (5.8) | 54 (5.9) | 0.973 |
| Dialysis, n (%) | 136 (86) | 45 (85) | 46 (85) | 0.994 |
| Transplantation, n (%) | 23 (14) | 8 (15) | 8 (15) | 0.991 |
| Mortality, n (%) | 474 (17) | 150 (16) | 153 (17) | 0.863 |
| **Follow-up Information** | | | | |
| Visits (times) | 8 (5, 13) | 8 (5, 13) | 8 (5, 13) | 0.314 |
| Observation time (y) | 5.55 (3.54, 7.37) | 5.60 (3.61, 7.42) | 5.62 (3.68, 7.50) | 0.574 |

Continuous variables are presented as the median (25th and 75th interquartile range). Categorical variables are presented as counts (n) and percentages (%).

## Appendix D.2. Baseline Characteristics by Kidney Failure

Tables D.2-D.5 presents baseline characteristics stratified by kidney failure status across the study cohorts. Patients who developed kidney failure generally had poorer baseline kidney function, particularly with respect to eGFR, albuminuria, blood pressure, and selected laboratory measures. However, the magnitude and direction of these differences varied across CK-NET-Yinzhou, PKUFH, C-STRIDE, and iCaReMe, reflecting cohort-specific heterogeneity.

Table D.2. Baseline Characteristics of the CK-NET-Yinzhou Development Cohort

| | **Overall (N = 4587)** | **With Kidney Failure (N = 266)** | **Without Kidney Failure (N = 4321)** | ***P* Value** |
|---|---|---|---|---|
| **Demographics** | | | | |
| Age (years) | 78 (69, 83) | 63 (58, 74) | 78 (69, 83) | <0.001*** |
| Male, n (%) | 2398 (52) | 164 (62) | 2234 (52) | 0.002** |
| History of Diabetes, n (%) | 1516 (33) | 92 (34) | 1424 (33) | 0.583 |
| Body Weight (kg) | 62.5 (56.1, 70.0) | 60.0 (55.0, 68.0) | 62.9 (56.0, 70.0) | 0.034* |
| Systolic BP (mmHg) | 152 (148, 160) | 155 (150, 160) | 152 (148, 160) | 0.018* |
| Diastolic BP (mmHg) | 90 (88, 98) | 90 (86, 98) | 90 (88, 98) | 0.417 |
| Mortality, n (%) | 777 (17) | 41 (15) | 736 (17) | 0.490 |
| **Clinical and Laboratory Characteristics** | | | | |
| eGFR (mL/min/1.73$m^2$) | 55 (44, 59) | 24 (14,47) | 55 (45,59) | <0.001*** |
| uACR (mg/g) | 11.9 (11.9, 25.2) | 202.0 (45.9, 411.4) | 16.32 (11.87, 39.21) | <0.001*** |
| Albumin (g/L) | 40.1 (36.4, 43.7) | 39.3 (35.6, 42.1) | 40.2 (36.5, 43.9) | <0.001*** |
| Serum Calcium (mmol/L) | 2.2 (2.1, 2.3) | 2.1 (2.0, 2.3) | 2.2 (2.1, 2.3) | <0.001*** |
| Serum Phosphorus (mmol/L) | 1.02 (0.91, 1.15) | 1.25 (1.09, 1.40) | 1.02 (0.90, 1.13) | <0.001*** |
| $HCO_3$ (mmol/L) | 24.1 (21.4, 27.0) | 21.0 (17.8, 23.5) | 24.5 (21.9, 27.0) | <0.001*** |
| Fasting Plasma Glucose (mmol/L) | 6.14 (5.33, 7.17) | 6.22 (5.36, 7.21) | 6.13 (5.30, 7.14) | 0.085 |
| Total Cholesterol (mmol/L) | 4.57 (3.79, 5.40) | 4.63 (4.01, 5.45) | 4.56 (3.74, 5.39) | 0.223 |
| LDL-C (mmol/L) | 2.55 (1.98, 3.16) | 2.58 (2.05, 3.18) | 2.54 (1.98, 3.16) | 0.571 |
| HDL-C (mmol/L) | 1.13 (0.95, 1.36) | 1.10 (0.93, 1.33) | 1.13 (0.95, 1.36) | 0.434 |
| Triglycerides (mmol/L) | 1.35 (0.94, 1.94) | 1.31 (0.90, 1.86) | 1.36 (0.94, 1.95) | 0.262 |
| **Follow-up Information** | | | | |
| Visits (times) | 8 (5,13) | 14 (6,30) | 8 (5,13) | <0.001*** |
| Observation time (y) | 5.57 (3.58,7.41) | 3.71 (2.35,6.17) | 4.25 (2.00,7.00) | 0.009** |

Continuous variables are presented as the median (25th and 75th interquartile range). Categorical variables are presented as counts (n) and percentages (%).

Table D.3. Baseline Characteristics of the PKUFH External Validation Cohort

| | **Overall (N = 934)** | **With Kidney Failure (N = 164)** | **Without Kidney Failure (N = 770)** | ***P* Value** |
|---|---|---|---|---|
| **Demographics** | | | | |
| Age (years) | 68 (57, 81) | 67 (57, 80) | 69 (57, 81) | 0.685 |
| Male, n (%) | 493 (53) | 85 (52) | 408 (53) | 0.787 |
| History of Diabetes, n (%) | 184 (20) | 24 (15) | 160 (21) | 0.072 |
| Body Weight (kg) | 64.8 (59.1, 72.5) | 64.2 (58.0, 71.8) | 65.0 (59.4, 72.7) | 0.324 |
| Systolic BP (mmHg) | 154 (136, 170) | 160 (140, 172) | 151 (130, 162) | <0.001*** |
| Diastolic BP (mmHg) | 92 (76, 107) | 93 (78, 110) | 92 (75, 104) | 0.098 |
| Mortality, n (%) | 14 (1.5) | 0 (0) | 14 (1.8) | <0.001*** |
| **Clinical and Laboratory Characteristics** | | | | |
| eGFR (mL/min/1.73$m^2$) | 33 (20, 47) | 16 (11, 24) | 35 (25, 50) | <0.001*** |
| uACR (mg/g) | 197.8 (25.2, 579.5) | 982.2 (371.1, 2045.1) | 155.0 (25.2, 468.4) | <0.001*** |
| Albumin (g/L) | 42.8 (39.9, 45.1) | 41.6 (38.7, 44.0) | 43.2 (40.7, 45.3) | <0.001*** |
| Serum Calcium (mmol/L) | 2.3 (2.2, 2.4) | 2.2 (2.1, 2.3) | 2.3 (2.2, 2.4) | <0.001*** |
| Serum Phosphorus (mmol/L) | 1.19 (1.06, 1.35) | 1.35 (1.18, 1.52) | 1.17 (1.05, 1.32) | <0.001*** |
| $HCO_3$ (mmol/L) | 24.4 (22.2, 26.4) | 22.3 (22.1, 22.8) | 25.0 (22.4, 27.0) | <0.001*** |
| Fasting Plasma Glucose (mmol/L) | 5.58 (4.99, 6.23) | 5.42 (4.86, 6.12) | 5.61 (5.01, 6.24) | 0.127 |
| Total Cholesterol (mmol/L) | 4.56 (3.83, 5.44) | 4.53 (3.75, 5.39) | 4.61 (3.85, 5.46) | 0.435 |
| LDL-C (mmol/L) | 2.41 (1.77, 2.93) | 2.38 (1.73, 2.88) | 2.42 (1.80, 2.94) | 0.628 |
| HDL-C (mmol/L) | 1.19 (1.00, 1.41) | 1.21 (1.04, 1.43) | 1.18 (0.99, 1.41) | 0.433 |
| Triglycerides (mmol/L) | 1.24 (0.87, 1.78) | 1.20 (0.85, 1.72) | 1.26 (0.88, 1.79) | 0.274 |
| **Follow-up Information** | | | | |
| Visits (times) | 11 (5, 23) | 11 (6, 22) | 10 (4, 23) | 0.323 |
| Observation time (y) | 3.33 (1.75, 5.75) | 1.92 (0.81, 3.27) | 2.50 (1.15, 6.25) | 0.003** |

Continuous variables are presented as the median (25th and 75th interquartile range). Categorical variables are presented as counts (n) and percentages (%).

Table D.4. Baseline Characteristics of the C-STRIDE External Validation Cohort.

| | **Overall (N = 1570)** | **With Kidney Failure (N = 459)** | **Without Kidney Failure (N = 1111)** | ***P* Value** |
|---|---|---|---|---|
| **Demographics** | | | | |
| Age (years) | 54 (43, 64) | 49 (40, 60) | 56 (45, 65) | <0.001*** |
| Male, n (%) | 886 (56) | 280 (61) | 606 (55) | 0.019* |
| History of Diabetes, n (%) | 308 (20) | 128 (28) | 180 (16) | <0.001*** |
| Body Weight (kg) | 65.0 (57.0, 74.0) | 64.0 (57.0, 74.0) | 65.0 (57.0, 74.0) | 0.205 |
| Systolic BP (mmHg) | 131 (121, 142) | 136 (124, 147) | 130 (120, 140) | <0.001*** |
| Diastolic BP (mmHg) | 80 (75, 88) | 83 (77, 90) | 80 (73, 87) | <0.001*** |
| Mortality, n (%) | 79 (5) | 29 (6) | 50 (5) | 0.134 |
| **Clinical and Laboratory Characteristics** | | | | |
| eGFR (mL/min/1.73$m^2$) | 32 (23, 43) | 24 (18, 32) | 36 (26, 46) | <0.001*** |
| uACR (mg/g) | 369.5 (69.4, 946.4) | 911.5 (361.6, 1839.6) | 253.8 (39.0, 630.4) | <0.001*** |
| Albumin (g/L) | 40.9 (36.7, 44.0) | 38.2 (33.3, 41.9) | 41.7 (38.6, 44.7) | <0.001*** |
| Serum Calcium (mmol/L) | 2.2 (2.1, 2.4) | 2.2 (2.0, 2.3) | 2.3 (2.1, 2.4) | <0.001*** |
| Serum Phosphorus (mmol/L) | 1.20 (1.06, 1.36) | 1.26 (1.11, 1.45) | 1.18 (1.05, 1.33) | <0.001*** |
| $HCO_3$ (mmol/L) | 24.7 (22.0, 27.5) | 23.3 (20.7, 26.6) | 25.0 (22.5, 27.6) | <0.001*** |
| Fasting Plasma Glucose (mmol/L) | 5.08 (4.54, 5.79) | 4.90 (4.41, 5.68) | 5.13 (4.60, 5.84) | <0.001*** |
| Total Cholesterol (mmol/L) | 4.64 (3.89, 5.43) | 4.56 (3.81, 5.43) | 4.68 (3.93, 5.43) | 0.291 |
| LDL-C (mmol/L) | 2.57 (2.06, 3.15) | 2.55 (2.06, 3.18) | 2.58 (2.06, 3.14) | 0.973 |
| HDL-C (mmol/L) | 1.06 (0.90, 1.32) | 1.05 (0.87, 1.29) | 1.07 (0.90, 1.33) | 0.255 |
| Triglycerides (mmol/L) | 1.77 (1.21, 2.44) | 1.78 (1.20, 2.39) | 1.75 (1.21, 2.45) | 0.879 |
| **Follow-up Information** | | | | |
| Visits (times) | 5 (3, 7) | 3 (2, 5) | 5 (4, 9) | <0.001*** |
| Observation time (y) | 2.43 (1.61, 3.70) | 2.01 (1.13, 3.07) | 2.43 (1.72, 3.75) | <0.001*** |

Continuous variables are presented as the median (25th and 75th interquartile range). Categorical variables are presented as counts (n) and percentages (%).

Table D.5. Baseline Characteristics of the iCaReMe External Validation Cohort.

| | **Overall (N = 498)** | **With Kidney Failure (N = 22)** | **Without Kidney Failure (N = 476)** | ***P* Value** |
|---|---|---|---|---|
| **Demographics** | | | | |
| Age (years) | 64 (55, 71) | 62 (54, 67) | 64 (55, 71) | 0.622 |
| Male, n (%) | 303 (61) | 13 (59) | 290 (61) | 0.863 |
| History of Diabetes, n (%) | 175 (35) | 13 (59) | 162 (34) | 0.016* |
| Body Weight (kg) | 65.0 (58.0, 75.0) | 66.0 (56.0, 75.0) | 65.0 (58.0, 75.0) | 0.849 |
| Systolic BP (mmHg) | 132 (121, 144) | 151 (145, 172) | 130 (121, 142) | <0.001*** |
| Diastolic BP (mmHg) | 80 (70, 86) | 80 (75, 88) | 80 (70, 86) | 0.638 |
| **Clinical and Laboratory Characteristics** | | | | |
| eGFR (mL/min/1.73$m^2$) | 38 (24, 49) | 13 (9, 19) | 38 (26, 50) | <0.001*** |
| uACR (mg/g) | 264.7 (70.2, 748.5) | 2251.5 (2143.3, 2432.4) | 143.9 (68.8, 721.6) | 0.005** |
| Albumin (g/L) | 41.7 (38.1, 44.2) | 34.1 (29.9, 39.7) | 42.0 (38.6, 44.2) | <0.001*** |
| Serum Calcium (mmol/L) | 2.3 (2.2, 2.4) | 2.2 (2.0, 2.2) | 2.3 (2.2, 2.4) | <0.001*** |
| Serum Phosphorus (mmol/L) | 1.21 (1.06, 1.38) | 1.41 (1.30, 1.63) | 1.19 (1.06, 1.37) | <0.001*** |
| $HCO_3$ (mmol/L) | 24.3 (21.6, 26.3) | 22.5 (19.7, 24.5) | 24.4 (21.8, 26.3) | 0.045* |
| Fasting Plasma Glucose (mmol/L) | 5.59 (5.00, 6.58) | 5.70 (5.20, 6.59) | 5.57 (4.99, 6.58) | 0.762 |
| Total Cholesterol (mmol/L) | 4.55 (3.91, 5.53) | 4.99 (4.35, 5.68) | 4.55 (3.88, 5.50) | 0.131 |
| LDL-C (mmol/L) | 2.59 (2.04, 3.24) | 3.11 (2.66, 3.52) | 2.56 (1.99, 3.20) | 0.039* |
| HDL-C (mmol/L) | 1.08 (0.90, 1.29) | 1.33 (0.96, 1.37) | 1.07 (0.90, 1.29) | 0.184 |
| Triglycerides (mmol/L) | 1.64 (1.17, 2.35) | 1.40 (1.07, 2.23) | 1.64 (1.20, 2.38) | 0.358 |
| **Follow-up Information** | | | | |
| Visits (times) | 3 (2, 3) | 3 (3, 3) | 3 (2, 3) | 0.249 |
| Observation time (y) | 1.00 (1.00, 1.00) | 0.46 (0.41, 0.56) | 1.00 (1.00, 1.00) | <0.001*** |

Continuous variables are presented as the median (25th and 75th interquartile range). Categorical variables are presented as counts (n) and percentages (%).

### Appendix D.3. Baseline Predictor Availability Before the Index Date

We examined whether each predictor had been recorded before the index date. Age and gender were available for all participants. Because cohort eligibility required at least two eGFR measurements ≤60 mL/min/1.73 m² separated by at least 3 months, with the second qualifying measurement defining the index date, every included participant had at least one eGFR measurement before the index date by design. Table D.6 summarizes the number and proportion of participants with at least one pre-index measurement for each of the remaining five longitudinal laboratory variables.

Table D.6. Availability of longitudinal laboratory variables before the index date.

| Dataset | Albumin | Calcium | Phosphorus | uACR | $HCO_3$ |
|---|---|---|---|---|---|
| CK-NET-Yinzhou | 4521(98.6%) | 4381(95.5%) | 4284(93.4%) | 4517(98.5%) | 1993(43.4%) |
| PKUFH | 830(88.9%) | 852(91.2%) | 848(90.8%) | 528(56.5%) | 170(18.2%) |
| C-STRIDE | 1552(98.9%) | 1555(99.0%) | 1551(98.8%) | 1570(100%) | 1520(96.8%) |
| iCaReMe | 428(85.9%) | 456(91.6%) | 430(86.3%) | 190(38.2%) | 301(60.4%) |

## Appendix E. Time Scale Chosen

The appropriate time scale has a great impact on modeling. That's because if the time scale is too short, more variables will be missing under the same timestamp, that is, there will be more missing values in Figure 1C(c); if the time scale is too long, more data will need to be merged, resulting in information loss. As a result, we compare three time scales: day, month, and year to balance the missing rate and merging rate of variables, and select month as the time scale.

The missing rates of variables at different time scales are shown in Table E.1. It can be seen that choosing day as the time scale results in an excessively high missing rate. However, choosing year as the time scale would significantly shorten the length of the time series, as Figure E.1 shows. Considering both the missing rate and the length of the time series, selecting month as the time scale is the most appropriate.

Table E.1: The Missing Rates of Variables at Different Time Scales.

| Time scale | eGFR | uACR | Albumin | Serum Calcium | Serum Phosphate | $HCO_3$ |
|---|---|---|---|---|---|---|
| **Day** | 41.4% | 57.8% | 51.0% | 43.6% | 51.2% | 40.6% |
| **Month** | 27.8% | 41.3% | 35.8% | 37.1% | 43.7% | 35.2% |
| **Year** | 20.0% | 32.2% | 23.8% | 29.6% | 35.8% | 33.8% |

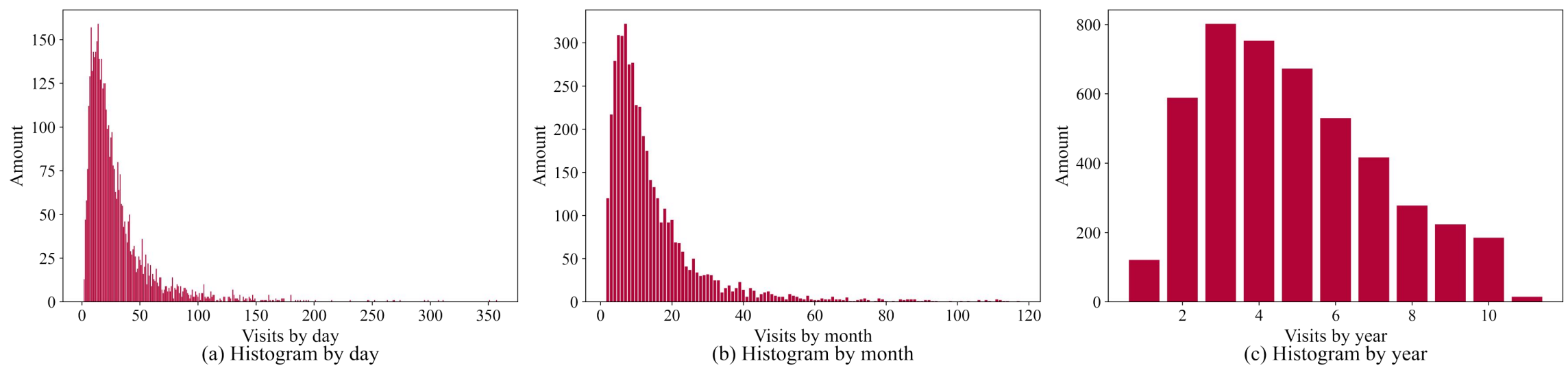

(a) Histogram by day (b) Histogram by month (c) Histogram by year

Figure E.1: Histogram of data length by different time scales

## Appendix F. Missing Data Imputation

### Appendix F.1. Missing Data Patterns

Before imputation, the availability and temporal spacing of the six longitudinal predictors were summarized as the median (interquartile range [IQR]) for each cohort (Tables F.1 and F.2). Measurement density therefore varied across cohorts and predictors. Importantly, the median intervals between consecutive observations were generally short, supporting the use of interpolation-based imputation for missing values within the monthly longitudinal sequences.

Table F.1: Distribution of the number of measurements per patient by longitudinal predictor and cohort before imputation.

| Cohorts | eGFR | uACR | Albumin | Serum Calcium | Serum Phosphate | $HCO_3$ |
|---|---|---|---|---|---|---|
| CK-NET-Yinzhou | 9(3,10) | 6(2,7) | 8(3,9) | 7(2,9) | 7(2,8) | 6(2,7) |
| PKUFH | 10(4,21) | 6(2,10) | 7(3,16) | 8(4,20) | 9(4,20) | 6(2,10) |
| C-STRIDE | 3(2,5) | 3(2,4) | 3(2,5) | 3(2,5) | 3(2,4) | 3(2,4) |
| iCaReMe | 3(2,3) | 2(1,3) | 2(1,3) | 2(1,3) | 2(1,3) | 2(1,3) |

Table F.2: Distribution of intervals, in months, between consecutive observations by longitudinal predictor and cohort before imputation.

| Cohorts | eGFR | uACR | Albumin | Serum Calcium | Serum Phosphate | $HCO_3$ |
|---|---|---|---|---|---|---|
| CK-NET-Yinzhou | 2(1,5) | 2(1,6) | 2(1,6) | 2(1,6) | 2(1,6) | 3(1,6) |
| PKUFH | 3(2,3) | 3(2,3) | 3(2,4) | 3(2,3) | 3(2,3) | 3(2,4) |
| C-STRIDE | 3(2,6) | 3(2,6) | 3(2,6) | 3(2,6) | 3(2,6) | 3(1,4) |
| iCaReMe | 6(6,6) | 6(6,6) | 6(6,6) | 6(6,6) | 6(6,6) | 6(6,6) |

### Appendix F.2. Missing Data Imputation Procedures

To address missing values in the longitudinal dataset, we implemented a structured three-step imputation procedure that leverages the temporal continuity of laboratory measurements. Let $x_t$ denote the observed value at time t, and $\hat{x}_t$ the imputed value.

1) **Forward and backward filling.** Forward filling is applied to all time points before the first non-missing observation, while backward filling is applied to all time points after the last non-missing observation, ensuring that abrupt discontinuities at the start or end of the observation period were avoided.

$$\hat{x}_t = \begin{cases} x_{t'} & t < t', if\ x_{t'} \text{ is the first non} - \text{missing value} \\ x_{t''} & t > t'', if\ x_{t''} \text{ is the last non} - \text{missing value} \end{cases}$$

2) **Linear interpolation.** Linear interpolation is applied to contiguous sequences of missing values located between two observed time points. For any missing time point $\mathrm{t}$ between $\mathrm{t}_1$ and $\mathrm{t}_2$, the imputed value is calculated as:

$$\hat{\mathrm{x}}_{\mathrm{t}} = \mathrm{x}_{\mathrm{t}_1} + \frac{(\mathrm{x}_{\mathrm{t}_2} - \mathrm{x}_{\mathrm{t}_1}) \cdot (\mathrm{t} - \mathrm{t}_1)}{\mathrm{t}_2 - \mathrm{t}_1},$$

where $\mathrm{t}_1$ and $\mathrm{t}_2$ are the nearest observed times bracketing the missing block, measured in months. This approach assigns progressively increasing or decreasing values across the gap.

3) **Median imputation.** Median imputation is reserved for the extreme case where a variable was missing for all time points in a given individual, replacing the entire series with the median value of that variable in the development dataset. Although no variable in the present development dataset was entirely missing, this step was included to ensure robustness in practical applications, where such cases may occur due to inconsistent testing schedules or incomplete data extraction from electronic health records. For instance, in the UK Biobank dataset, the HCO_3 indicator was not collected at all; in this case, we imputed all $HCO_3$ values using the median from the development dataset.

To illustrate how this rule would apply in practice, we consider a representative patient's monthly laboratory monitoring data, comprising six longitudinal variables and their corresponding months.

$$\{Time, eGFR, Albumin, Serum\ Calcium(Ca), Serum\ Phosphate(P), uACR, HCO_3\}$$

Table F.3 below shows the observed data, with blank entries indicating missing values.

Table F.3: The original dataset with missing values.

| **Time (yyyymm)** | **eGFR** | **albumin** | **Ca** | **P** | **uACR** | $HCO_3$ |
|---|---|---|---|---|---|---|
| 2010/01 | | 44.1 | 2.4 | 1.29 | 337.4104 | |
| 2010/04 | 31.372689 | 39.95 | | 1.306667 | 229.07014 | 28.0 |
| 2010/05 | 29.878915 | 39.65 | 2.245 | 1.225 | 201.98507 | |
| 2010/06 | 28.889378 | 44.5 | 2.43 | 1.065 | | 29.5 |
| 2011/01 | 30.084284 | 44.3 | 2.27 | 1.29 | 66.559747 | 31.0 |

| 2011/02 | 32.055332 | 45.6 | 2.29 | 1.27 | 337.4104 | 29.6 |
|---|---|---|---|---|---|---|
| 2011/07 | 27.958122 | 43.9 | 2.31 | 1.38 | | |

The first step is to identify values that require forward fill or backward fill. Forward fill is used for all months before the first non-missing observation for a variable, while backward fill is used for all months after the last non-missing observation. This ensures temporal continuity at both edges of the time series. In Table F.4, cells highlighted in blue indicate missing values to be replaced via forward fill, and cells in orange indicate those to be replaced via backward fill.

Table F.4: Identification of missing values requiring forward or backward filling

| **Time (yyyymm)** | **eGFR** | **albumin** | **Ca** | **P** | **uACR** | $HCO_3$ |
|---|---|---|---|---|---|---|
| 2010/01 | | 44.1 | 2.40 | 1.29 | 337.41 | |
| 2010/04 | 31.37 | 39.9 | | 1.31 | 229.07 | 28.0 |
| 2010/05 | 29.88 | 39.6 | 2.25 | 1.22 | 201.99 | |
| 2010/06 | 28.89 | 44.5 | 2.43 | 1.07 | | 29.5 |
| 2011/01 | 30.08 | 44.3 | 2.27 | 1.29 | 66.56 | 31.0 |
| 2011/02 | 32.06 | 45.6 | 2.29 | 1.27 | 337.41 | 29.6 |
| 2011/07 | 27.96 | 43.9 | 2.31 | 1.38 | | |

After applying these replacements, Table F.5 presents the dataset following forward and backward filling. The boundary gaps are now filled, but some missing values remain within the observation window. These internal gaps will be handled in the next step.

Table F.5: Values after forward/backward filling

| **Time (yyyymm)** | **eGFR** | **albumin** | **Ca** | **P** | **uACR** | $HCO_3$ |
|---|---|---|---|---|---|---|
| 2010/01 | 31.37 | 44.1 | 2.40 | 1.29 | 337.41 | 28.0 |
| 2010/04 | 31.37 | 39.9 | | 1.31 | 229.07 | 28.0 |
| 2010/05 | 29.88 | 39.6 | 2.25 | 1.22 | 201.99 | |
| 2010/06 | 28.89 | 44.5 | 2.43 | 1.07 | | 29.5 |
| 2011/01 | 30.08 | 44.3 | 2.27 | 1.29 | 66.56 | 31.0 |
| 2011/02 | 32.06 | 45.6 | 2.29 | 1.27 | 337.41 | 29.6 |
| 2011/07 | 27.96 | 43.9 | 2.31 | 1.38 | 337.41 | 29.6 |

We then identify missing values suitable for linear interpolation. In Table F.6, these values are highlighted in green.

Table F.6: Identification of missing values requiring linear interpolation

| **Time (yyyymm)** | **eGFR** | **albumin** | **Ca** | **P** | **uACR** | $HCO_3$ |
|---|---|---|---|---|---|---|
| 2010/01 | 31.37 | 44.1 | 2.40 | 1.29 | 337.41 | 28.0 |
| 2010/04 | 31.37 | 39.9 | | 1.31 | 229.07 | 28.0 |
| 2010/05 | 29.88 | 39.6 | 2.25 | 1.22 | 201.99 | |
| 2010/06 | 28.89 | 44.5 | 2.43 | 1.07 | | 29.5 |

| 2011/01 | 30.08 | 44.3 | 2.27 | 1.29 | 66.56 | 31.0 |
|---|---|---|---|---|---|---|
| 2011/02 | 32.06 | 45.6 | 2.29 | 1.27 | 337.41 | 29.6 |
| 2011/07 | 27.96 | 43.9 | 2.31 | 1.38 | 337.41 | 29.6 |

For each gap, the two nearest observed values before and after the gap are used to compute intermediate values.

To illustrate the linear interpolation process, we use the Calcium (Ca) variable in a representative example. Two observed values are available:

$$t_1 = 2010/01, x_{t_1} = 2.40 mmol/L,$$

$$t_2 = 2010/05, x_{t_2} = 2.25 mmol/L,$$

The value at 2010/04 is missing and will be imputed using the linear interpolation formula:

$$\hat{x}_t = x_{t_1} + \frac{(x_{t_2} - x_{t_1}) \cdot (t - t_1)}{t_2 - t_1},$$

Substituting the observed values into the interpolation formula yields:

$$\begin{aligned} \hat{x}_{2010/04} &= 2.40 + \frac{(2.25 - 2.40) \times (2010/04 - 2010/01)}{(2010/05 - 2010/01)} \\ &= 2.40 + \frac{(-0.15) \times (3\text{months})}{(4\text{months})} \\ &= 2.40 - 0.01125 \\ &= 2.2875 \text{ mmol/L} \end{aligned}$$

Thus, the missing Ca value for 2010/04 is imputed as 2.29 mmol/L (rounded to two decimal places). The same approach can be applied to compute the imputed values for other variables and time points.

Finally, Table F.7 shows the completed dataset after linear interpolation. All variables now form uninterrupted time series, which can be directly fed into the predictive model.

Table F.7: Final dataset after linear interpolation

| **Time (yyyymm)** | **eGFR** | **albumin** | **Ca** | **P** | **uACR** | $HCO_3$ |
|---|---|---|---|---|---|---|
| 2010/01 | 31.37 | 44.1 | 2.40 | 1.29 | 337.41 | 28.0 |
| 2010/04 | 31.37 | 39.9 | 2.29 | 1.31 | 229.07 | 28.0 |
| 2010/05 | 29.88 | 39.6 | 2.25 | 1.22 | 201.99 | 28.8 |
| 2010/06 | 28.89 | 44.5 | 2.43 | 1.07 | 185.06 | 29.5 |
| 2011/01 | 30.08 | 44.3 | 2.27 | 1.29 | 66.56 | 31.0 |
| 2011/02 | 32.06 | 45.6 | 2.29 | 1.27 | 337.41 | 29.6 |

| 2011/07 | 27.96 | 43.9 | 2.31 | 1.38 | 337.41 | 29.6 |
|---|---|---|---|---|---|---|

This step-by-step approach provides full transparency for how missing laboratory values are addressed in our longitudinal dataset, ensuring that imputation is conducted systematically and reproducibly.

## Appendix G. Baseline Models

The baseline comparators were grouped according to how they use information available at a prediction time. The static Kidney Failure Risk Equations (KFREs) use one set of demographic and laboratory measurements at the index date [6]. The dynamic Cox model updates risk using the latest measurements available during follow-up [23]. The shared-parameter joint model uses the complete longitudinal eGFR history observed up to the prediction time to estimate the underlying kidney-function trajectory and link that trajectory to the subsequent risk of kidney failure [25].

We used the non-North America calibrated Kidney Failure Risk Equations (KFREs), including the four-variable and eight-variable 2-year models [6]. As our dynamic prediction model focuses on short-term risk prediction, the 2-year KFREs were selected as the closest established benchmark. These models generate a fixed-horizon risk estimate from a single reference measurement. For each predictor, we used the observed value closest to the index date within a six-month window. The index date was defined as the date of the second qualifying eGFR measurement $\leq 60$ mL/min/1.73 $m^2$ in the longitudinal cohorts and the first qualifying measurement in UK Biobank.

$$
\begin{aligned}
\text{KFRE 4-variable 2-year probability} = 1 - 0.9832{*}{*}\exp(&-0.2201 \times (\text{age}/10 - 7.036) \\
&+ 0.2467 \times (\text{male} - 0.5642) \\
&- 0.5567 \times (\text{eGFR}/5 - 7.222) \\
&+ 0.4510 \times (\log(\text{ACR}) - 5.137)),
\end{aligned}
$$

$$
\begin{aligned}
\text{KFRE 8-variable 2-year probability} = 1 - 0.9827{*}{*}\exp(&-0.1992 \times (\text{age}/10 - 7.036) \\
&+ 0.1602 \times (\text{male} - 0.5642) \\
&- 0.4919 \times (\text{eGFR}/5 - 7.222) \\
&+ 0.3364 \times (\log(\text{ACR}) - 5.137)
\end{aligned}
$$

$$
\begin{aligned}
&- 0.3441 \times (\text{albumin} - 3.997) \\
&+ 0.2604 \times (\text{phosphate} - 3.916) \\
&- 0.07354 \times (\text{HCO3} - 25.57) \\
&- 0.2228 \times (\text{calcium} - 9.355)),
\end{aligned}
$$

The dynamic Cox model was adapted from the latest-available-measurement model described by Tangri et al. [23]. It uses a Cox proportional hazards framework in which sex and uACR are retained as baseline covariates, whereas age, eGFR, serum albumin, phosphate, calcium, and bicarbonate are treated as time-updated covariates. At each eligible visit, a new risk estimate is produced using only information observed on or before that visit. When a laboratory variable is not newly measured, its most recent available value is carried forward until it is updated. Thus, the covariate history is represented as a sequence of piecewise-constant values, and the estimated hazard changes when updated clinical information becomes available.

The shared-parameter joint model was adapted from the approach described by van den Brand et al. [25]. It simultaneously fits two linked components. First, a longitudinal mixed-effects submodel uses all eGFR measurements available up to the prediction time to estimate each patient's underlying eGFR trajectory. Fixed effects describe the average trajectory associated with follow-up time and baseline characteristics, while patient-specific random effects allow the level and rate of eGFR change to differ between individuals. Second, a survival submodel relates the subsequent kidney-failure hazard to age, sex, baseline uACR, and the estimated current eGFR level and slope. The two submodels are connected through shared patient-specific parameters. As a result, longitudinal eGFR and time to kidney failure are estimated jointly rather than as independent processes. The current eGFR level and slope used by the survival component are latent, model-based quantities informed by all prior eGFR observations, rather than a single potentially noisy measurement. This formulation accounts for measurement error and between-patient heterogeneity in kidney-function trajectories and allows the longitudinal biomarker process to be associated with the event process.

For both dynamic baseline models, a 1-year prediction horizon was used to match the proposed dynamic model. As risk estimates can be updated whenever new clinical information becomes available, these models naturally support repeated short-term prediction [25].

## Appendix H. Additional Evaluation Metrics

Beyond AUROC, we evaluated model performance using metrics that are informative under class imbalance, including precision-recall (PR) analysis, threshold-dependent classification metrics, and probability-based calibration metrics. Comparisons included the four- and eight-variable static KFREs, the Latest-measurement Cox model, the shared-parameter joint model, and KFDeep, where applicable.

Tables H.1-H.4 summarise the quantitative performance metrics derived from the PR curves and ROC curves for the four cohorts. For each model, we calculated average precision (AP) as the area under the PR curve, and derived sensitivity (recall), specificity, positive predictive value (PPV), negative predictive value (NPV), F1-score, and balanced accuracy (ACC) from the confusion matrix at the optimal decision threshold. The optimal threshold for each model was determined using the Youden index (maximising sensitivity + specificity – 1), a commonly used criterion for balancing true positive and true negative rates. Calibration performance was assessed using the Brier score and expected calibration error (ECE) with 10-bin partitioning, based on predicted probabilities and observed event rates.

Table H.1: Performance comparison on internal CK-NET-Yinzhou test dataset.

| Metric | KFRE 4v2y | KFRE 8v2y | Latest-measurement Cox model | Shared-parameter joint model | KFDeep |
|---|---|---|---|---|---|
| **Prediction Horizon** | 2-year (prevalence = 0.05) | | 1-year (prevalence = 0.06) | | |
| **Sensitivity** (↑) | 0.7872 | 0.8085 | 0.8519 | 0.8333 | **0.8889** |
| **Specificity** (↑) | 0.8381 | 0.8416 | 0.8345 | 0.8449 | **0.9097** |
| **PPV** (↑) | 0.2079 | 0.2159 | 0.2434 | 0.2514 | **0.3810** |
| **NPV** (↑) | 0.9865 | 0.9879 | 0.9890 | 0.9878 | **0.9924** |
| **AP** (↑) | 0.5062 | 0.5463 | 0.5543 | 0.5117 | **0.6755** |
| **F1-Score** (↑) | 0.3289 | 0.3408 | 0.3786 | 0.3863 | **0.5333** |
| **Balanced ACC** (↑) | 0.8127 | 0.8250 | 0.8432 | 0.8391 | **0.8993** |
| **Brier Score** (↓) | 0.0441 | 0.0426 | 0.0399 | 0.0437 | **0.0323** |
| **Scaled Brier Score** (↑) | 0.0922 | 0.1230 | 0.2793 | 0.2107 | **0.4166** |
| **ECE (10 bins)** (↓) | 0.0205 | 0.0191 | 0.0259 | 0.0263 | **0.0070** |

The best-performing result is shown in bold, and the second-best result is underlined.

Table H.2: Performance comparison on external PKUFH dataset.

| Metric | KFRE 4v2y | KFRE 8v2y | Latest-measurement Cox model | Shared-parameter joint model | KFDeep |
|---|---|---|---|---|---|
| **Prediction Horizon** | 2-year (prevalence = 0.17) | | 1-year (prevalence = 0.17) | | |
| **Sensitivity** (↑) | 0.8194 | 0.8323 | 0.8333 | 0.8415 | **0.8476** |
| **Specificity** (↑) | 0.5725 | 0.5751 | 0.5959 | 0.6364 | **0.6779** |
| **PPV** (↑) | 0.2761 | 0.2804 | 0.3020 | 0.3301 | **0.3592** |
| **NPV** (↑) | 0.9409 | 0.9451 | 0.9446 | 0.9496 | **0.9543** |
| **AP** (↑) | 0.2801 | 0.2835 | 0.3011 | 0.3794 | **0.4794** |
| **F1-Score** (↑) | 0.4130 | 0.4195 | 0.4433 | 0.4742 | **0.5045** |
| **Balanced ACC** (↑) | 0.6959 | 0.7037 | 0.7146 | 0.7389 | **0.7627** |
| **Brier Score** (↓) | 0.1418 | 0.1372 | 0.1415 | 0.1398 | **0.1256** |
| **Scaled Brier Score** (↑) | 0.0109 | 0.0430 | 0.0225 | 0.0342 | **0.1323** |
| **ECE (10 bins)** (↓) | 0.1301 | 0.1184 | 0.1053 | 0.0911 | **0.0640** |

The best-performing result is shown in bold, and the second-best result is underlined.

Table H.3: Performance comparison on external C-STRIDE dataset.

| **Metric** | **KFRE 4v2y** | **KFRE 8v2y** | **Latest-measurement Cox model** | **Shared-parameter joint model** | **KFDeep** |
|---|---|---|---|---|---|
| **Prediction Horizon** | 2-year (prevalence = 0.15) | | 1-year (prevalence = 0.29) | | |
| **Sensitivity** (↑) | 0.6840 | 0.6883 | 0.8039 | 0.7647 | **0.8214** |
| **Specificity** (↑) | 0.7984 | 0.7640 | 0.6481 | 0.6661 | **0.6823** |
| **PPV** (↑) | 0.3692 | 0.3347 | 0.4855 | 0.4861 | **0.5164** |
| **NPV** (↑) | **0.9361** | 0.9342 | 0.8889 | 0.8726 | 0.9024 |
| **AP** (↑) | 0.4058 | 0.4504 | 0.6352 | 0.5972 | **0.6911** |
| **F1-Score** (↑) | 0.4504 | 0.4795 | 0.6054 | 0.5944 | **0.6341** |
| **Balanced ACC** (↑) | 0.7262 | 0.7412 | 0.7260 | 0.7154 | **0.7518** |
| **Brier Score** (↓) | 0.1217 | **0.1197** | 0.1788 | 0.1803 | 0.1569 |
| **Scaled Brier Score** (↑) | 0.0302 | 0.0461 | 0.1357 | 0.1285 | **0.2416** |
| **ECE (10 bins)** (↓) | 0.1192 | 0.1084 | 0.1097 | 0.1209 | **0.0787** |

The best-performing result is shown in bold, and the second-best result is underlined.

Table H.4: Performance comparison on external iCaReMe dataset.

| **Metric** | **Latest-measurement Cox model** | **Shared-parameter joint model** | **KFDeep** |
|---|---|---|---|
| **Prediction Horizon** | 1-year (prevalence = 0.04) | | |
| **Sensitivity** (↑) | 0.9091 | **0.9545** | **0.9545** |
| **Specificity** (↑) | 0.8151 | 0.7983 | **0.8235** |
| **PPV** (↑) | 0.1852 | 0.1795 | **0.2000** |
| **NPV** (↑) | 0.9949 | 0.9974 | **0.9975** |
| **AP** (↑) | 0.4021 | 0.4669 | **0.5462** |
| **F1-Score** (↑) | 0.3077 | 0.3022 | **0.3307** |
| **Balanced ACC** (↑) | 0.8621 | 0.8764 | **0.8890** |
| **Brier Score** (↓) | 0.0374 | 0.0356 | **0.0313** |
| **Scaled Brier Score** (↑) | 0.1143 | 0.1569 | **0.2587** |
| **ECE (10 bins)** (↓) | 0.0647 | 0.0569 | **0.0561** |

The best-performing result is shown in bold, and the second-best result is underlined.

Across the four longitudinal cohorts, KFDeep consistently achieved the highest AP, reaching 0.6755 in CK-NET-Yinzhou, 0.4794 in PKUFH, 0.6911 in C-STRIDE, and 0.5462 in iCaReMe, compared with 0.5543, 0.3794, 0.6352, and 0.4669, respectively, for the strongest baseline in each cohort. KFDeep also achieved the highest F1-score and balanced accuracy in all four cohorts, together with the highest or tied sensitivity and generally favourable specificity, PPV, and NPV among the same-horizon dynamic comparators. Overall, these findings indicate that KFDeep's advantage extends beyond AUROC and is accompanied by generally favourable threshold-based performance and probability accuracy.

In addition to the quantitative results, we further analysed the graphical results to illustrate model behaviour. Figure H.1 presents the PR curves for KFDeep and the applicable baseline models in the internal CK-NET-Yinzhou, PKUFH, C-STRIDE, and iCaReMe cohorts. The operating point selected using the Youden index is marked on each curve. Consistent with the tabulated AP values, KFDeep maintained a generally more favourable precision-recall trade-off and achieved the highest AP in each cohort.

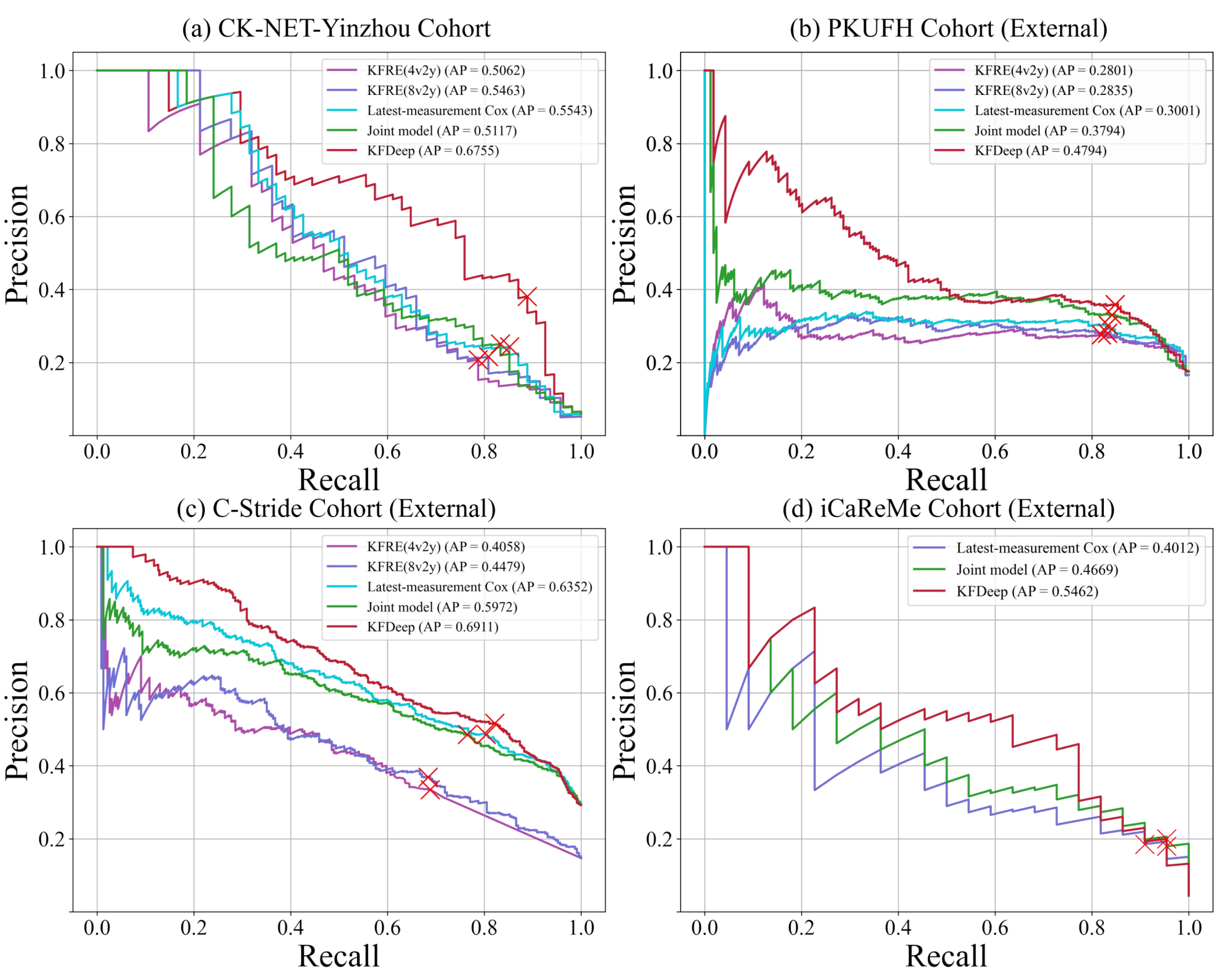

Figure H.1: Precision-Recall Curves.

Figure H.2 presents the complete 10-bin calibration curves for KFDeep across the four cohorts, together with Wilson 95% confidence intervals and the distribution of predicted risks. Predicted and empirical risks were generally aligned, although deviations and confidence intervals became larger in sparsely populated risk bins, particularly in cohorts with fewer outcome events.

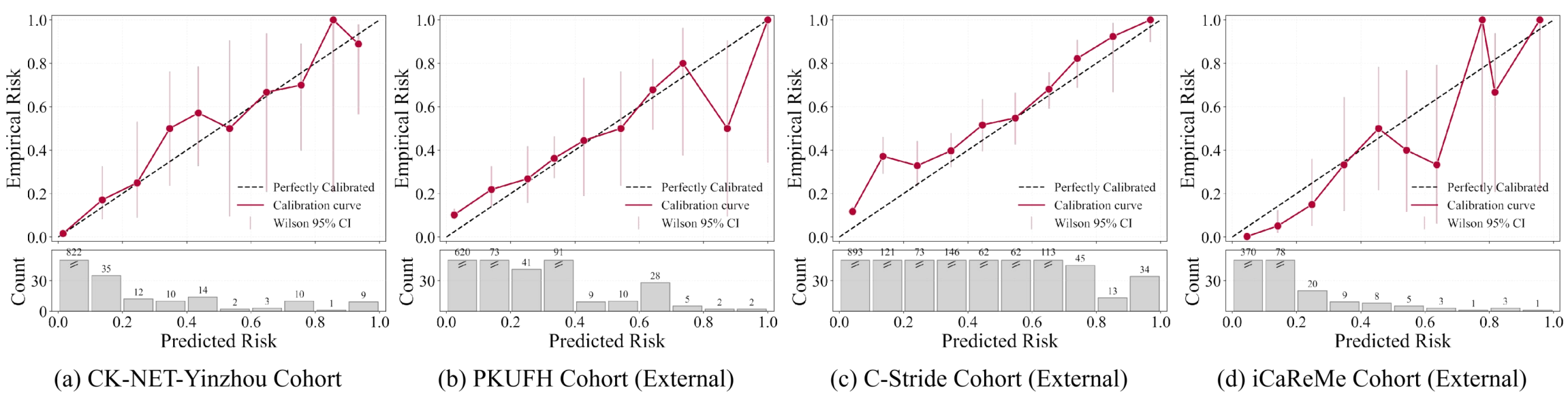

(a) CK-NET-Yinzhou Cohort (b) PKUFH Cohort (External) (c) C-Stride Cohort (External) (d) iCaReMe Cohort (External)

Figure H.2: 10-bin Calibration Curves.

## Appendix I. Pairwise DeLong Tests for AUC Comparison

To statistically compare AUROC values between models, we conducted pairwise DeLong tests for the four longitudinal datasets (Tables I.1-I.4). Results with $p < 0.05$ were considered statistically significant. In the tables, *, **, and *** indicate $p < 0.05$, $p < 0.01$, and $p < 0.001$, respectively.

Across the four longitudinal cohorts, KFDeep achieved significantly higher AUROCs than all applicable static and dynamic baseline models in CK-NET-Yinzhou, PKUFH, and C-STRIDE. In iCaReMe, KFDeep had a numerically higher AUROC than dynamic Cox and the joint model, but neither difference reached statistical significance; static KFREs were not evaluated because the available follow-up did not support their established 2-year prediction horizon. No statistically significant AUROC differences were observed among the applicable baseline models within any of the four cohorts. Overall, the DeLong analyses support a consistent discrimination advantage for KFDeep.

Table I.1: Pairwise DeLong test p-values between models for the internal CK-NET-Yinzhou test dataset.

| | **KFRE-4v2y** | **KFRE-8v2y** | **Dynamic Cox** | **Joint Model** | **KFDeep** |
|---|---|---|---|---|---|
| **KFRE-4v2y** | – | 0.1776 | 0.3928 | 0.3398 | 0.0052** |
| **KFRE-8v2y** | | – | 0.6451 | 0.5656 | 0.0098** |
| **Dynamic Cox** | | | – | 0.7348 | 0.0467* |
| **Joint Model** | | | | – | 0.0424* |
| **KFDeep** | | | | | – |

Table I.2: Pairwise DeLong test p-values between models for the external PKUFH dataset.

| | KFRE-4v2y | KFRE-8v2y | Dynamic Cox | Joint Model | KFDeep |
|---|---|---|---|---|---|
| **KFRE-4v2y** | – | 0.1446 | 0.2167 | 0.0669 | 0.0075** |
| **KFRE-8v2y** | | – | 0.6444 | 0.2886 | 0.0203* |
| **Dynamic Cox** | | | – | 0.5518 | 0.0349* |
| **Joint Model** | | | | – | 0.0485* |
| **KFDeep** | | | | | – |

Table I.3: Pairwise DeLong test p-values between models for the external C-STRIDE dataset.

| | KFRE-4v2y | KFRE-8v2y | Dynamic Cox | Joint Model | KFDeep |
|---|---|---|---|---|---|
| **KFRE-4v2y** | – | 0.1975 | 0.0588 | 0.2539 | <0.001*** |
| **KFRE-8v2y** | | – | 0.5251 | 0.8827 | 0.0054** |
| **Dynamic Cox** | | | – | 0.3765 | 0.0107* |
| **Joint Model** | | | | – | <0.001*** |
| **KFDeep** | | | | | – |

Table I.4: Pairwise DeLong test p-values between models for the external iCaReMe dataset.

| | Dynamic Cox | Joint Model | KFDeep |
|---|---|---|---|
| **Dynamic Cox** | – | 0.5498 | 0.0947 |
| **Joint Model** | | – | 0.1144 |
| **KFDeep** | | | – |

## Appendix J. Dynamic Comparative Experiments

Tables J.1-J.3 report the detailed AUROC scores for KFDeep and the dynamic baseline models using the first 1-15 cumulative visits, with a 1-year prediction horizons, on the CK-NET-Yinzhou, PKUFH, and C-STRIDE validation datasets, which had sufficient repeated measurements for this analysis. These results correspond to the temporal evaluation framework described in the Methods section, where models were evaluated based on their ability to predict kidney failure using longitudinal information accumulated up to a given visit. Pairwise DeLong tests were conducted between KFDeep and each dynamic baseline at every visit count. Significance levels are denoted as follows: * for $P < 0.05$, ** for $P < 0.01$, and *** for $P < 0.001$.

In CK-NET-Yinzhou, KFDeep had a higher AUROC than both baselines at all 15 visit counts, with statistically significant differences at 11 visit counts versus each baseline. In PKUFH, KFDeep likewise had the highest AUROC at every visit count and significantly outperformed the dynamic Cox model at 9 of 15 visit counts and the joint model at 7 of 15. In C-STRIDE, KFDeep had a higher AUROC than the joint model at all visit counts and exceeded the dynamic Cox model at 14 of 15 visit counts; the corresponding differences were significant at 13 and 11 visit counts, respectively. Despite visit-level fluctuations, KFDeep showed a significant positive association between cumulative visit count and AUROC in all three cohorts (CK-NET-Yinzhou: Spearman's $\rho = 0.62$, $p = 0.014$; PKUFH: $\rho = 0.79$, $p < 0.001$; C-STRIDE: $\rho = 0.70$, $p = 0.003$). By comparison, no significant trend was observed for the dynamic Cox model, while the joint model showed significant positive trends in PKUFH and C-STRIDE but not in CK-NET-Yinzhou. These results indicate that KFDeep generally benefited from accumulating longitudinal information.

Table J.1. AUROC scores based on cumulative visits for the internal CK-NET-Yinzhou test dataset.

| Visit Count | Dynamic Cox | *P* Value (vs KFDeep) | Joint Model | *P* Value (vs KFDeep) | KFDeep |
|---|---|---|---|---|---|
| 1 | 0.8471 | 0.046* | 0.8290 | 0.008** | 0.9103 |
| 2 | 0.8076 | 0.151 | 0.7845 | 0.018* | 0.8448 |
| 3 | 0.8214 | 0.002** | 0.8190 | 0.001** | 0.9110 |
| 4 | 0.7963 | 0.021* | 0.8464 | 0.526 | 0.8663 |
| 5 | 0.8316 | 0.063 | 0.7247 | <0.001*** | 0.8845 |
| 6 | 0.7977 | 0.013* | 0.8237 | 0.085 | 0.8688 |
| 7 | 0.7602 | <0.001*** | 0.8631 | 0.007** | 0.9420 |
| 8 | 0.8708 | 0.007** | 0.8001 | <0.001*** | 0.9477 |
| 9 | 0.8980 | 0.030* | 0.7856 | <0.001*** | 0.9533 |
| 10 | 0.8182 | <0.001*** | 0.8142 | <0.001*** | 0.9206 |
| 11 | 0.8704 | 0.509 | 0.7768 | <0.001*** | 0.8892 |
| 12 | 0.7923 | <0.001*** | 0.8811 | 0.457 | 0.9044 |
| 13 | 0.9131 | 0.234 | 0.8893 | 0.047* | 0.9491 |
| 14 | 0.8641 | 0.008** | 0.8916 | 0.130 | 0.9420 |
| 15 | 0.8855 | 0.002** | 0.9064 | 0.031* | 0.9698 |

Table J.2. AUROC scores based on cumulative visits for the external PKUFH dataset.

| Visit Count | Dynamic Cox | *P* Value (vs KFDeep) | Joint Model | *P* Value (vs KFDeep) | KFDeep |
|---|---|---|---|---|---|
| 1 | 0.7456 | 0.069 | 0.7667 | 0.247 | 0.7882 |
| 2 | 0.7627 | 0.240 | 0.7702 | 0.438 | 0.7752 |
| 3 | 0.7395 | 0.027* | 0.7527 | 0.085 | 0.7843 |

| 4 | 0.7442 | 0.087 | 0.7524 | 0.355 | 0.7675 |
|---|---|---|---|---|---|
| 5 | 0.7160 | 0.016* | 0.7311 | 0.028* | 0.7738 |
| 6 | 0.7698 | 0.121 | 0.7482 | 0.012* | 0.7801 |
| 7 | 0.7320 | 0.015* | 0.7399 | 0.036* | 0.7906 |
| 8 | 0.7475 | 0.037* | 0.7577 | 0.043* | 0.7814 |
| 9 | 0.7371 | 0.002** | 0.7273 | <0.001*** | 0.8290 |
| 10 | 0.7532 | 0.006** | 0.7728 | 0.028* | 0.8259 |
| 11 | 0.7626 | 0.122 | 0.7735 | 0.153 | 0.7989 |
| 12 | 0.7793 | 0.094 | 0.7884 | 0.107 | 0.8019 |
| 13 | 0.7556 | 0.021* | 0.7963 | 0.776 | 0.8056 |
| 14 | 0.7730 | 0.045* | 0.7854 | 0.064 | 0.8148 |
| 15 | 0.7690 | 0.009* | 0.7946 | 0.013** | 0.8377 |

Table J.3. AUROC scores based on cumulative visits for the external C-STRIDE dataset.

| Visit Count | Dynamic Cox | *P* Value (vs KFDeep) | Joint Model | *P* Value (vs KFDeep) | KFDeep |
|---|---|---|---|---|---|
| 1 | 0.8053 | 0.024* | 0.7987 | 0.005** | 0.8469 |
| 2 | 0.8075 | 0.105 | 0.7864 | 0.009** | 0.8298 |
| 3 | 0.8184 | 0.118 | 0.8281 | 0.278 | 0.8393 |
| 4 | 0.8279 | 0.217 | 0.7976 | 0.135 | 0.8159 |
| 5 | 0.7940 | 0.042* | 0.7613 | 0.003** | 0.8217 |
| 6 | 0.7831 | 0.015* | 0.7801 | 0.012* | 0.8236 |
| 7 | 0.7655 | 0.136 | 0.7421 | 0.016* | 0.7864 |
| 8 | 0.8059 | 0.003** | 0.7890 | <0.001*** | 0.8630 |
| 9 | 0.7646 | 0.002** | 0.8031 | 0.009** | 0.8505 |
| 10 | 0.7836 | 0.008** | 0.7783 | 0.005** | 0.8263 |
| 11 | 0.8001 | 0.006** | 0.8138 | 0.011* | 0.8575 |
| 12 | 0.7986 | <0.001*** | 0.8285 | 0.006** | 0.8646 |
| 13 | 0.8474 | <0.001*** | 0.8609 | 0.002** | 0.9165 |
| 14 | 0.8663 | <0.001*** | 0.8485 | <0.001*** | 0.9093 |
| 15 | 0.8822 | <0.001*** | 0.9255 | 0.044* | 0.9556 |

## Appendix K. Single-Timepoint External Validation on the UK Biobank Cohort

To further assess the generalizability of the KFDeep model across different ethnic backgrounds, we conducted an additional exploratory external validation using the UK Biobank (UKBB), a large, prospective cohort from the United Kingdom. Predominantly comprising European participants, UKBB represents a distinct healthcare context from our primary datasets. Our original longitudinal EHR setting requires multi-visit laboratory data, which are challenging to obtain in diverse populations, and the UKBB provides a rare alternative with large-scale, high-quality, standardized data that enables evaluation in a demographically and clinically different population. Although UKBB contains only single-timepoint laboratory measurements and does not include bicarbonate ($HCO_3^-$), its scale, diversity, and rigorous data collection make it a valuable testbed for examining cross-population performance. This analysis is presented in the appendix to complement our primary longitudinal evaluations.

### Appendix K.1. Study Population

The selection process for the UKBB analysis cohort is shown in Figure K.1. We began with all UKBB participants who had a baseline serum creatinine measurement. Estimated glomerular filtration rate (eGFR) was calculated using the CKD-EPI equation, and individuals with an eGFR $\leq$ 60 mL/min/1.73 m² at baseline were retained. To further ensure the inclusion of participants with CKD stage 3–5, we required at least one qualifying ICD-10 code from hospital or primary care records: N183 (CKD stage 3), N184 (CKD stage 4), N185 (CKD stage 5), N189 (CKD unspecified), I120 (hypertensive CKD with stage 5 CKD or ESRD), or I132 (hypertensive heart and CKD with stage 5 CKD or ESRD). Participants with missing values for any of the predictor variables required for the analysis were then excluded. Finally, individuals with a recorded kidney failure event prior to or at the baseline visit were removed. After these steps, 3,261 participants remained and constituted the study population for external validation.

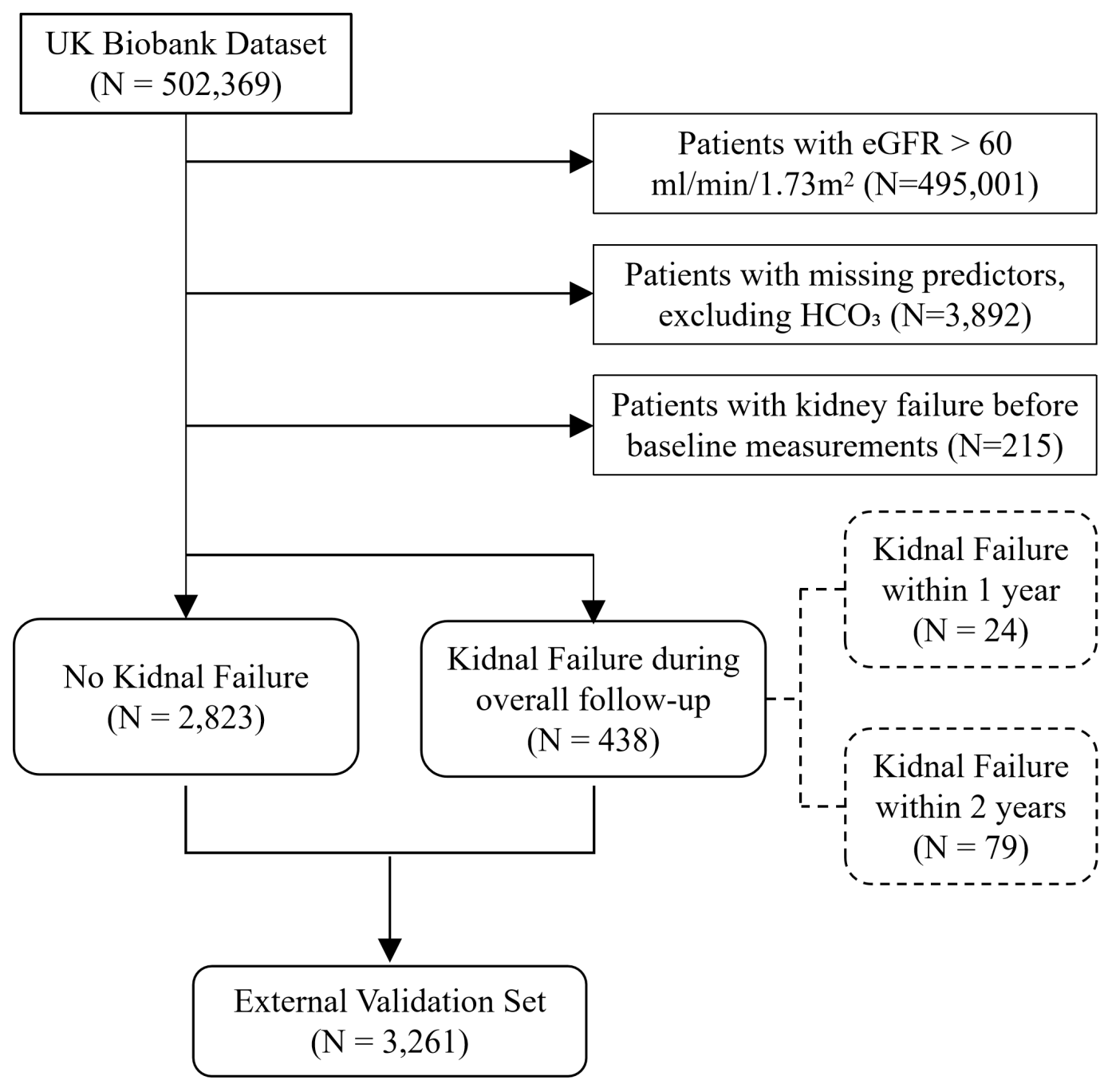

Figure K.1: Flowchart of Study Enrollment and Inclusion Criteria in UKBB Dataset.

## Appendix K.2. Predictors and Outcomes

The predictors included the eight standard variables used in the Kidney Failure Risk Equations (KFREs): six laboratory variables—estimated glomerular filtration rate (eGFR), urine albumin-to-creatinine ratio (uACR), albumin, serum calcium, serum phosphate, and bicarbonate ($HCO_3$)—and two demographic variables: age and gender. eGFR was calculated from serum creatinine using the CKD-EPI equation. In UKBB, all laboratory predictors were measured at a single baseline visit. Besides, $HCO_3$ was unavailable in UKBB and was imputed using the same development-cohort median for all participants in models requiring it (KFRE-8v2y, latest-measurement Cox, and KFDeep). This ensured consistent handling without introducing artificial between-participant variation, but the absence of observed $HCO_3$ remains an important limitation. KFRE-4v2y and the joint model do not use $HCO_3$.

The outcome was initiation of kidney replacement therapy (KRT), defined as the start of maintenance dialysis (hemodialysis or peritoneal dialysis) or kidney transplantation. KRT events were identified from hospitalization and outpatient records using either: (1) Diagnosis codes indicating maintenance dialysis, kidney transplantation, or dialysis access care: Z992, Z940, Z490, Z491, Z492, T861, T824, Y841 or (2) Procedure codes indicating dialysis or kidney transplantation: X404, X403, X402, X401, X411, X412, M012, M013. Consistent with the analyses in the other cohorts, the established KFRE-4v2y and KFRE-8v2y models were evaluated for 2-year kidney failure, whereas the Latest-measurement Cox model, shared-parameter joint model, and KFDeep were evaluated for 1-year kidney failure. Because the latter models received only baseline measurements in UKBB, their predictions should be regarded as single-timepoint applications of otherwise dynamic models.

### Appendix K.3. Results

Continuous variables were summarized as medians with interquartile ranges (IQRs), and compared between groups using the Mann-Whitney U test. Categorical variables were expressed as counts (n) and percentages (%) and compared using the chi-square test or Fisher's exact test, as appropriate. All p-values were two-sided, and a significance threshold of 0.05 was applied. Statistical significance levels are denoted as $p < 0.05$ (*), $p < 0.01$ (**), and $p < 0.001$ (***). Baseline characteristics are presented in Table K.1 in the format of median (IQR) for continuous variables and n (%) for categorical variables.

Table K.1: Baseline Characteristics of the UK Biobank Study Population, Stratified by Kidney Failure Status During Follow-up

| | Overall (N = 3261) | With Kidney Failure (N = 438) | Without Kidney Failure (N = 2823) | P Value |
|---|---|---|---|---|
| **Demographics** | | | | |
| Age (years) | 65 (60,67) | 62 (56,66) | 65 (61,68) | <0.001*** |
| Male, n (%) | 1790 (54.9) | 295 (67.4) | 1495 (53.0) | <0.001*** |
| Country, n (%) | | | | 0.3511 |
| England | 2870 (88.0) | 393 (89.7) | 2477 (87.7) | |
| Scotland | 246 (7.5) | 31 (7.1) | 215 (7.1) | |
| Wales | 145 (4.5) | 14 (3.2) | 131 (4.6) | |
| History of Diabetes, n (%) | 734 (22.5) | 140 (32.0) | 594 (21.0) | <0.001*** |
| Body Weight (kg) | 83.60 (73.00,95.20) | 84.20 (72.70,97.55) | 83.50 (73.00,94.83) | 0.2036 |
| Systolic BP (mmHg) | 141.00 (128.00,155.00) | 143.00 (130.75,154.50) | 140.50 (127.50,155.00) | 0.0909 |
| Diastolic BP (mmHg) | 80.50 (73.50,88.50) | 82.00 (73.50,88.25) | 80.50 (73.50,88.50) | 0.6739 |
| Mortality, n (%) | 2195 (67.3) | 218 (49.8) | 1977 (70.0) | <0.001*** |
| **Clinical and Laboratory Characteristics** | | | | |
| eGFR (mL/min/1.73m2 ) | 52.85 (45.60,57.06) | 39.43 (28.30,49.68) | 53.76 (47.94,57.36) | <0.001*** |
| uACR (mg/g) | 17.17 (8.00,64.21) | 202.17 (40.86,819.41) | 14.26 (7.29,40.52) | <0.001*** |
| Albumin (g/L) | 44.10 (42.16,46.11) | 42.74 (40.58,44.83) | 44.30 (42.42,46.26) | <0.001*** |
| Serum Calcium (mmol/L) | 2.38 (2.31,2.45) | 2.34 (2.26,2.42) | 2.38 (2.31,2.45) | <0.001*** |
| Serum Phosphorus (mmol/L) | 1.17 (1.05,1.29) | 1.19 (1.07,1.33) | 1.16 (1.05,1.28) | 0.0005*** |
| Fasting Plasma Glucose (mmol/L) | 5.14 (4.70,5.84) | 5.15 (4.68,6.16) | 5.14 (4.70,5.79) | 0.1992 |
| Total Cholesterol (mmol/L) | 4.99 (4.21,5.96) | 4.72 (3.97,5.61) | 5.04 (4.23,6.00) | <0.001*** |
| LDL-C (mmol/L) | 3.03 (2.47,3.77) | 2.87 (2.36,3.59) | 3.06 (2.49,3.80) | 0.0001*** |
| HDL-C (mmol/L) | 1.21 (1.01,1.48) | 1.14 (0.95,1.39) | 1.22 (1.02,1.49) | <0.001*** |
| Triglycerides (mmol/L) | 1.85 (1.31,2.63) | 1.91 (1.37,2.77) | 1.84 (1.31,2.61) | 0.3432 |
| **Follow-up Information** | | | | |
| Visits (times) | 1 (1,1) | 1 (1,1) | 1 (1,1) | 1.0000 |
| Observation time (y) | 13.76 (13.04,14.38) | 13.64 (12.98,14.35) | 13.77 (13.06,14.39) | 0.0673 |

Continuous variables are presented as the median (25th and 75th interquartile range). Categorical variables are presented as counts (n) and percentages (%).

The final study population comprised 3,261 participants, of whom 438 developed kidney failure during follow-up and 2,823 did not. Compared with the non-KF group, the KF group was younger, had a higher proportion of males, lower eGFR, higher uACR, lower albumin, lower serum calcium, and higher serum phosphate, consistent with a high-risk CKD phenotype. UKBB participants contributed only a single baseline visit, and the median follow-up duration was 13.7 years. The 438 events reflect the entire available follow-up and should be distinguished from the horizon-specific event prevalence used for model evaluation, which was 79 for the 2-year KFRE outcomes and 24 for the 1-year dynamic-model outcomes.

Table K.2 summarizes model performance in UKBB. The two KFREs predicted 2-year kidney failure, whereas the dynamic Cox model, joint model, and KFDeep predicted 1-year kidney failure. Among the 1-year models, KFDeep achieved the highest AUROC (0.8830), AP (0.4963), and balanced accuracy (0.8684), together with the lowest Brier score (0.0055) and ECE (0.0047). The 2-year KFREs achieved AUROCs of 0.8815 and 0.8709, although their results are not directly comparable with those of the 1-year models because the prediction horizons differed. Notably, the 8-variable KFRE did not outperform the 4-variable KFRE as expected. This likely reflects the absence of $HCO_3$ measurements in UKBB. Pairwise DeLong tests identified no statistically significant AUROC differences between the evaluated models (Table K.3).

Table K.2: Performance comparison on external UKBB dataset.

| **Metric** | **KFRE 4v2y** | **KFRE 8v2y** | **Latest-measurement Cox model** | **Shared-parameter joint model** | **KFDeep** |
|---|---|---|---|---|---|
| **Prediction Horizon** | 2-year (prevalence = 0.02) | | 1-year (prevalence = 0.007) | | |
| **AUROC (↑)** | 0.8815 (0.8456, 0.9174) | 0.8709 (0.8362, 0.9056) | 0.8407 (0.7715, 0.9099) | 0.8279 (0.7540, 0.9018) | **0.8830** (0.8128, 0.9532) |
| **Sensitivity (↑)** | **0.7975** | 0.7848 | 0.6250 | 0.5833 | 0.7500 |
| **Specificity (↑)** | 0.8454 | 0.8727 | 0.8860 | 0.9462 | **0.9867** |
| **PPV (↑)** | 0.1135 | 0.1328 | 0.0391 | 0.0745 | **0.2951** |
| **NPV (↑)** | 0.9941 | 0.9939 | 0.9969 | 0.9967 | **0.9981** |
| **AP (↑)** | 0.4243 | 0.2635 | 0.2737 | 0.2543 | **0.4963** |
| **F1-Score (↑)** | 0.1987 | 0.2271 | 0.0735 | 0.1321 | **0.4235** |
| **Balanced ACC (↑)** | 0.8214 | 0.8288 | 0.7555 | 0.7648 | **0.8684** |
| **Brier Score (↓)** | 0.0142 | 0.0180 | 0.0065 | 0.0068 | **0.0055** |
| **Scaled Brier Score (↑)** | **0.3993** | 0.2385 | 0.1103 | 0.0692 | 0.2471 |
| **ECE (10 bins) (↓)** | 0.0076 | 0.0104 | 0.0053 | 0.0056 | **0.0047** |

The best-performing result is shown in bold, and the second-best result is underlined.

Table K.3: Pairwise DeLong test p-values between models for the external UKBB dataset.

| | **KFRE-4v2y** | **KFRE-8v2y** | **Dynamic Cox** | **Joint Model** | **KFDeep** |
|---|---|---|---|---|---|
| **KFRE-4v2y** | – | 0.2225 | 0.2545 | 0.2364 | 0.7141 |
| **KFRE-8v2y** | | – | 0.2576 | 0.2392 | 0.7221 |
| **Dynamic Cox** | | | – | 0.1928 | 0.3855 |
| **Joint Model** | | | | – | 0.3258 |
| **KFDeep** | | | | | – |

Figure K.2 shows the PR curves for all models and the calibration and decision-curve analyses for KFDeep. KFDeep achieved the highest AP, showed generally reasonable calibration despite uncertainty in sparsely populated risk groups, and provided positive net benefit across a range of threshold probabilities.

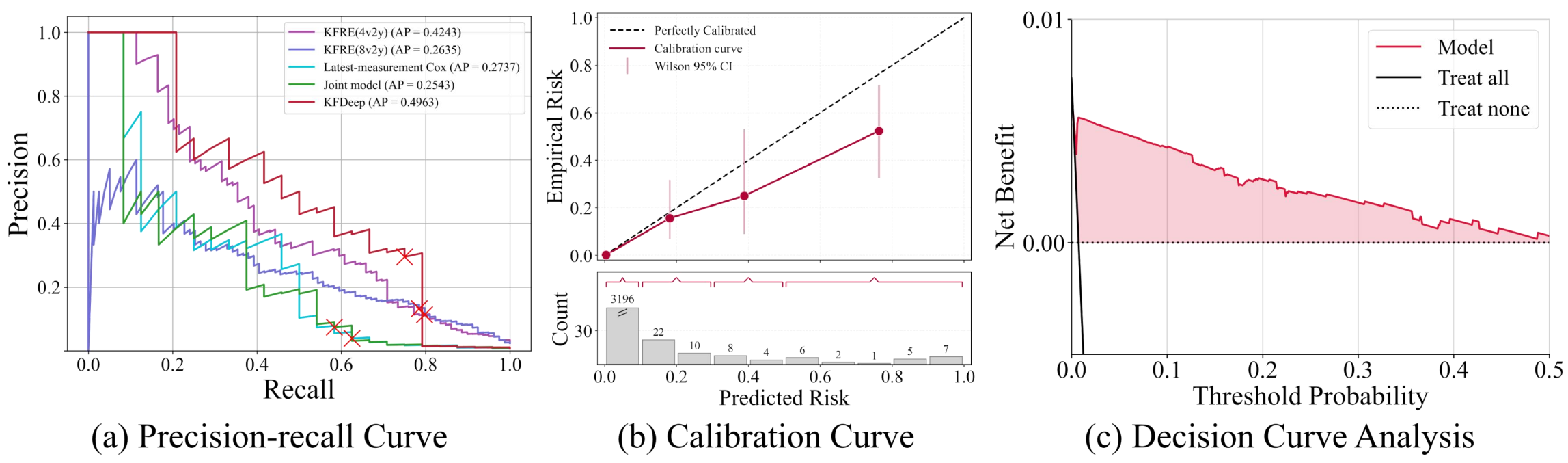

(a) Precision-recall Curve (b) Calibration Curve (c) Decision Curve Analysis

Figure K.2: Model performance in the UKBB external dataset.

## Appendix L. Results on Cross-Validation

In the 5-fold cross-validation analysis, patients were first divided into five disjoint folds at the patient level to avoid data leakage across training and evaluation sets. Each fold contained a similar number of kidney failure (KF) cases, with 54 cases assigned to fold 1 and 53 cases in each of the remaining folds. In each iteration, KFDeep was trained from scratch on three folds, validated on a fourth fold, and tested on the remaining fold, with these roles rotated across the five iterations. In the main text, we reported results from fold 1 for consistency with the original experimental split, while Appendix L presents results for all folds. The model architecture and training configuration were prespecified and were not selected by comparing performance across folds. This analysis was therefore intended as a sensitivity analysis of performance stability across independent data partitions.

Across the five folds, the model achieved consistently high discrimination and good calibration, as summarized in Table L.1, which reports all performance metrics. Specificity and sensitivity remained well balanced across folds, and no statistically significant AUROC differences were observed in the pairwise DeLong tests, as shown in Table L.2. Figure L.1 presents the precision–recall curves, and Figure L.2 shows the calibration curves using both 5-bin and 10-bin groupings; these visual assessments were also highly consistent across folds.

The overall trends, both quantitative and visual, indicate robust model performance across different patient splits. These results confirm that KFDeep's performance is not dependent on a specific data split and that its design generalizes well across multiple, independently constructed training-validation-test partitions.

Table L.1: Model performance metrics across five folds of cross-validation.

| Metric | fold1 | fold2 | fold3 | fold4 | Fold5 | Mean ± Std |
|---|---|---|---|---|---|---|
| AUROC (↑) (95% CI) | 0.9311 (0.8873,0.9749) | 0.9414 (0.9093,0.9735) | 0.9223 (0.8714,0.9732) | **0.9530** (0.9244,0.9816) | 0.9366 (0.8950,0.9782) | 0.9369 ± 0.0115 |
| AP (↑) | 0.6755 | 0.6571 | 0.6701 | **0.7013** | 0.6791 | 0.6766 ± 0.0161 |
| Sensitivity (↑) | **0.8889** | 0.8679 | 0.8491 | 0.8868 | 0.8679 | 0.8721 ± 0.0163 |
| Specificity (↑) | 0.9097 | 0.9282 | **0.9572** | 0.8993 | 0.9456 | 0.9280 ± 0.0241 |
| PPV (↑) | 0.3810 | 0.4259 | **0.5488** | 0.3507 | 0.4946 | 0.4402 ± 0.0814 |
| NPV (↑) | 0.9924 | 0.9913 | 0.9904 | **0.9923** | 0.9915 | 0.9916 ± 0.0008 |
| F1-Score (↑) | 0.5333 | 0.5714 | **0.6667** | 0.5027 | 0.6301 | 0.5808 ± 0.0676 |
| Balanced ACC | 0.8993 | 0.8981 | 0.9031 | 0.8930 | **0.9068** | 0.9001 ± 0.0052 |
| Brier Score (↓) | 0.0323 | 0.0311 | 0.0290 | 0.0289 | **0.0285** | 0.0300 ± 0.0017 |
| ECE (10 bins) | **0.0070** | 0.0259 | 0.0209 | 0.0161 | 0.0229 | 0.0186 ± 0.0074 |
| ECE (5 bins) | 0.0064 | 0.0187 | **0.0061** | 0.0123 | 0.0126 | 0.0112 ± 0.0052 |

Table L.2: Pairwise DeLong test p-values across five-fold cross-validation.

| CV Split | CV-1 | CV-2 | CV-3 | CV-4 | CV-5 |
|---|---|---|---|---|---|
| **CV-1** | - | 0.7157 | 0.7995 | 0.4078 | 0.8619 |
| **CV-2** | | - | 0.5458 | 0.6039 | 0.8052 |
| **CV-3** | | | - | 0.3062 | 0.6797 |
| **CV-4** | | | | - | 0.5353 |
| **CV-5** | | | | | - |

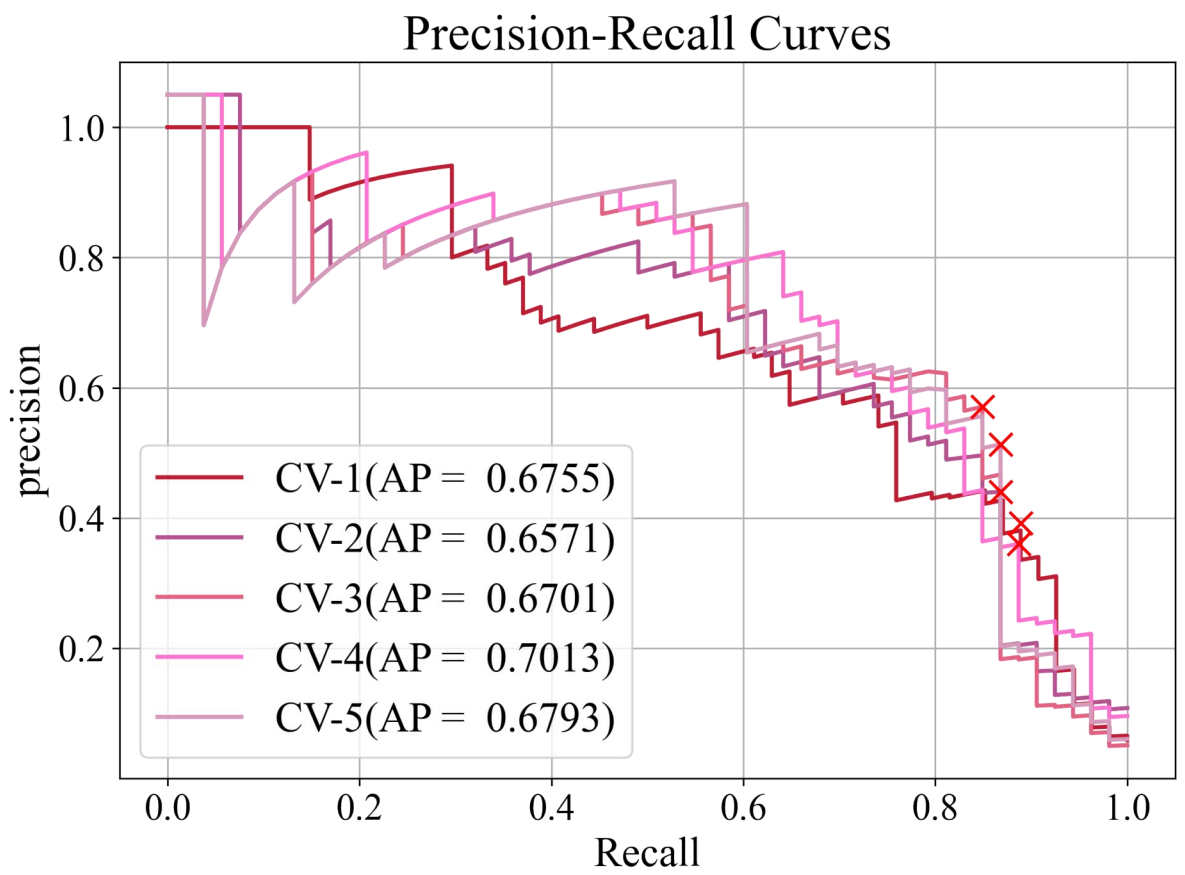

Figure L.1: Precision-Recall Curves across 5-fold Cross-Validation.

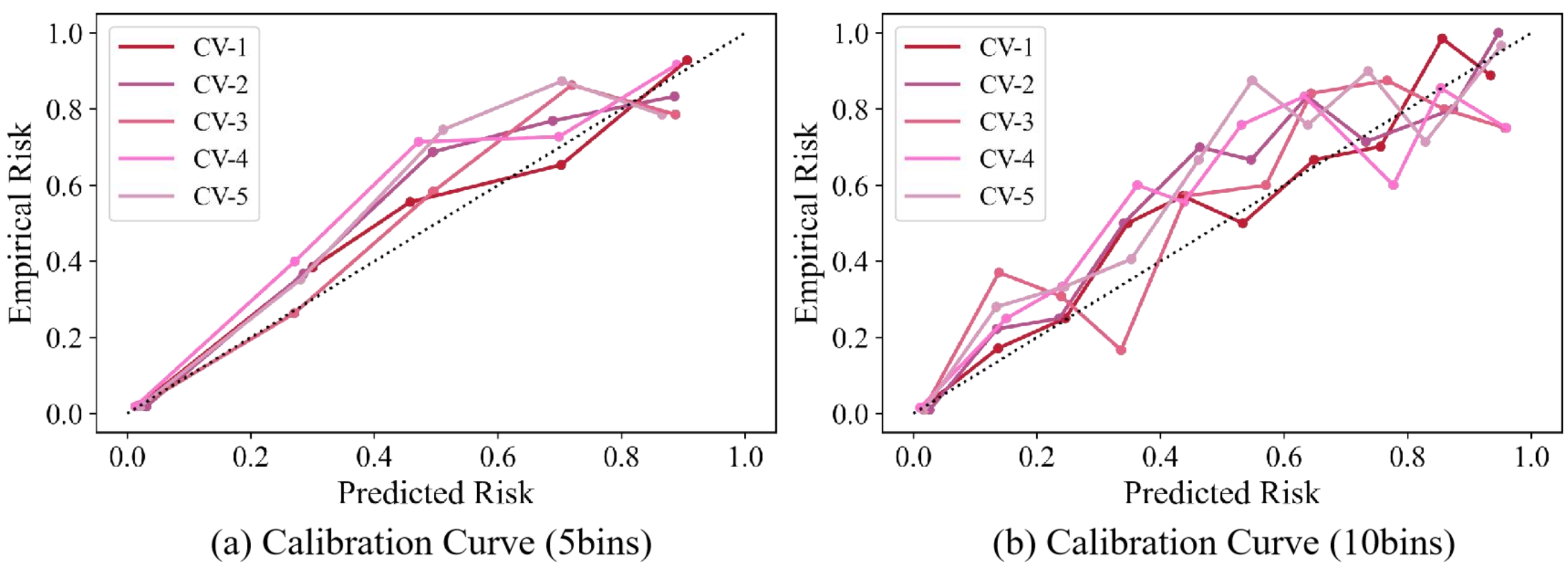

(a) Calibration Curve (5bins)

(b) Calibration Curve (10bins)

Figure L.2: Calibration Curves Across 5-Fold Cross-Validation.

## Appendix M. Model Details

The KFDeep model can calculate the risk ratio of patients with CKD for kidney failure based on patients' electronic health records. The patient's EHR data is represented as follows:

$$\mathrm{EHR}=[\mathrm{EHR}_{t_0},\ldots,\mathrm{EHR}_{t_n}]^\top ,$$

where the $\mathrm{EHR}_{t_n}$ can be represented as:

$$\mathrm{EHR}_{t_i}=[\Delta t_{t_i}, \mathrm{eGFR}_{t_i}, \mathrm{Albumin}_{t_i}, \mathrm{Blood\ Calcium}_{t_i}, \mathrm{Blood\ Phosphorus}_{t_i}, \mathrm{uACR}_{t_i}, \mathrm{HCO3}_{t_i}]^\top ,$$

where $\Delta t$ represents the time interval between patient assessments. When i=0, the $\Delta t_{t_0}=0$; when i>0, the $\Delta t_{t_i}$ is the difference in months between the occurrence time of $EHR_{t_i}$ and the occurrence time of $EHR_{t_{i-1}}$.

Feed $EHR_{t_0}$ through $EHR_{t_n}$ sequentially into the following formula for iterative calculation. At each moment t, the calculation formula can be represented as:

$$C^S_{t-1}=\tanh(W_d C_{t-1}+b_d),$$

$$g(\Delta t_{t_i})=\frac{1}{\Delta t_{t_i}+0.00001},$$

$$\widehat{C}^S_{t-1}=C^S_{t-1}*g(\Delta t_{t_i}),$$

$$C^T_{t-1}=C_{t-1}-C^S_{t-1},$$

$$C^*_{t-1}=C^T_{t-1}+\widehat{C}^S_{t-1},$$

$$f_t=\sigma(W_f EHR_t+U_f h_{t-1}+b_f),$$

$$i_t=\sigma(W_i EHR_t+U_i h_{t-1}+b_i),$$

$$o_t=\sigma(W_o EHR_t+U_o h_{t-1}+b_o),$$

$$\widetilde{C}=\tanh(W_c EHR_t+U_c h_{t-1}+b_c),$$

$$C_t=f_t*C^*_{t-1}+i_t*\widetilde{C},$$

$$h_t=o_t*\tanh(C_t).$$

The $W_d$, $b_d$, $W_f$, $U_f$, $b_f$, $W_i$, $U_i$, $b_i$, $W_o$, $U_o$, $b_o$, $W_c$, $U_c$, $b_c$ are all constant matrices trained by the neural network. When t=0, both $C_0$ and $h_0$ are initialized as zero matrices. The tanh and $\sigma$ functions used in the above formula are as follows:

$$\tanh(x)=\frac{e^x-e^{-x}}{e^x+e^{-x}},$$

$$\sigma(x)=\frac{1}{1+e^{-x}}.$$

The obtained $C_t$, $h_t$, and $EHR_{t+1}$ are then used together as inputs for the next time step, until $h_n$ is calculated.

$$h_b = \text{ReLU}(f(W_1 h_n + b_1)),$$

where $h_b$ is the final embedding, and ReLU is an activation function defined as:

$$\text{ReLU}(x) = \max(0, x).$$

Then, concatenate the patient's static variables (gender and age) with $h_b$ to obtain the $h_a$ variable:

$$h_a = [h_b^\top, \text{age}, \text{gender}]^\top,$$

where $h_a$ is input into the following formula to obtain the final prediction:

$$\text{prediction} = \sigma(f(W_2 h_a + b_2)).$$

Similarly, $W_1$, $b_1$, $W_2$, $b_2$ are also constant matrices obtained by neural network training. The final prediction is the model prediction.

Finally, we calibrated the model prediction by quantiles to obtain risk values between 0 and 1. For example, if the risk is 0.6, it indicates that the individual has a higher risk of kidney failure than 60% of the population.

## Appendix N. Model Implementation

We implemented a single-layer time-aware LSTM model with a hidden dimension of 16. The model was trained with a learning rate of $1 \times 10^{-2}$, a batch size of 16, and a total of 20 epochs. The loss function used was binary cross-entropy, and model optimization was performed using the Adam optimizer. All experiments were conducted using PyTorch. Additional implementation details are provided in Table N.1.

Table N.1: Model Implementation Settings

| Parameter | Value |
| --- | --- |
| Model architecture | Time-aware LSTM (hidden size = 16) followed by a 2-layer MLP:<br>Layer 1: 16 → 6 with ReLU<br>Layer 2: 8 → 1 with Sigmoid |
| Learning rate | $1 \times 10^{-2}$ |
| Batch size | 16 |
| Number of epochs | 20 |
| Loss function | Binary Cross-Entropy |
| Optimizer | Adam |
| LR scheduler | ReduceLROnPlateau |
| Scheduler settings | mode = min, factor = 0.8, patience = 10 |
| Framework | PyTorch |
| Python version | 3.83 |
| Hardware | CPU |

## Appendix O. Time-aware Ablation Study

To isolate the contribution of irregular time-interval modeling, we compared KFDeep with a conventional LSTM using the same predictors and evaluation pipeline. The conventional LSTM was therefore treated as an architectural ablation rather than a clinical baseline.
KFDeep achieved higher performance than the LSTM in all four longitudinal cohorts (Table O.1). The corresponding PR curves are shown in Figure O.1. The smaller AUROC difference in iCaReMe, where measurements followed a more regular 6-month schedule, is consistent with the time-aware mechanism being particularly useful when observation intervals are irregular.

This distinction is relevant to real-world clinical data, in which laboratory tests are obtained at irregular intervals and identical sequences of measurements may have different implications depending on the elapsed time between visits. By incorporating these intervals, KFDeep can adjust the contribution of previous information rather than treating observations as equally spaced. When only a single measurement is available, no inter-visit interval exists and the time-aware adjustment has no previous temporal state to modify; KFDeep therefore reduces to the same single-step temporal computation as a conventional LSTM.

Table O.1: Performance comparison between KFDeep and a conventional LSTM without explicit time interval modeling across four cohorts.

| **Method** | **LSTM** | **KFDeep** | **LSTM** | **KFDeep** |
|---|---|---|---|---|
| **Dataset** | CK-NET-Yinzhou (prevalence = 0.06) | | PKUFH (prevalence = 0.17) | |
| **AUROC (↑)** | 0.9015 (0.8535, 0.9495) | **0.9311** (0.8873, 0.9749) | 0.7709 (0.7238, 0.8180) | **0.8141** (0.7728, 0.8554) |
| **Sensitivity (↑)** | 0.8704 | **0.8889** | 0.8344 | **0.8476** |
| **Specificity (↑)** | 0.8472 | **0.9097** | 0.6070 | **0.6779** |
| **PPV (↑)** | 0.2626 | **0.3810** | 0.3098 | **0.3592** |
| **NPV (↑)** | 0.9905 | **0.9924** | 0.9455 | **0.9543** |
| **AP (↑)** | 0.5800 | **0.6755** | 0.4082 | **0.4794** |
| **F1-Score (↑)** | 0.4034 | **0.5333** | 0.4518 | **0.5045** |

| **Balanced ACC** (↑) | 0.8588 | **0.8993** | 0.7207 | **0.7627** |
|---|---|---|---|---|
| **Brier Score** (↓) | 0.0356 | **0.0323** | 0.1321 | **0.1256** |
| **Scaled Brier Score** (↑) | 0.3570 | **0.4166** | 0.0874 | **0.1323** |
| **ECE (10 bins)** (↓) | 0.0122 | **0.0070** | 0.0865 | **0.0640** |
| **Dataset** | C-STRIDE (prevalence = 0.29) | | iCaReMe (prevalence = 0.04) | |
| **AUROC** (↑) | 0.8136 (0.7904, 0.8358) | 0.8427 (0.8209, 0.8645) | 0.9294 (0.9032, 0.9521) | 0.9359 (0.9024, 0.9694) |
| **Sensitivity** (↑) | 0.8126 | **0.8214** | **1.0000** | 0.9545 |
| **Specificity** (↑) | 0.6634 | **0.6823** | 0.8193 | **0.8235** |
| **PPV** (↑) | 0.4993 | **0.5164** | **0.2037** | 0.2000 |
| **NPV** (↑) | 0.8955 | **0.9024** | **1.0000** | 0.9975 |
| **AP** (↑) | 0.6527 | **0.6911** | 0.4920 | **0.5462** |
| **F1-Score** (↑) | 0.6186 | **0.6341** | 0.3077 | **0.3307** |
| **Balanced ACC** (↑) | 0.7380 | **0.7518** | 0.8621 | **0.8890** |
| **Brier Score** (↓) | 0.1631 | **0.1569** | 0.0341 | **0.0313** |
| **Scaled Brier Score** (↑) | 0.2116 | **0.2416** | 0.1924 | **0.2587** |
| **ECE (10 bins)** (↓) | 0.0912 | **0.0787** | 0.0580 | **0.0561** |

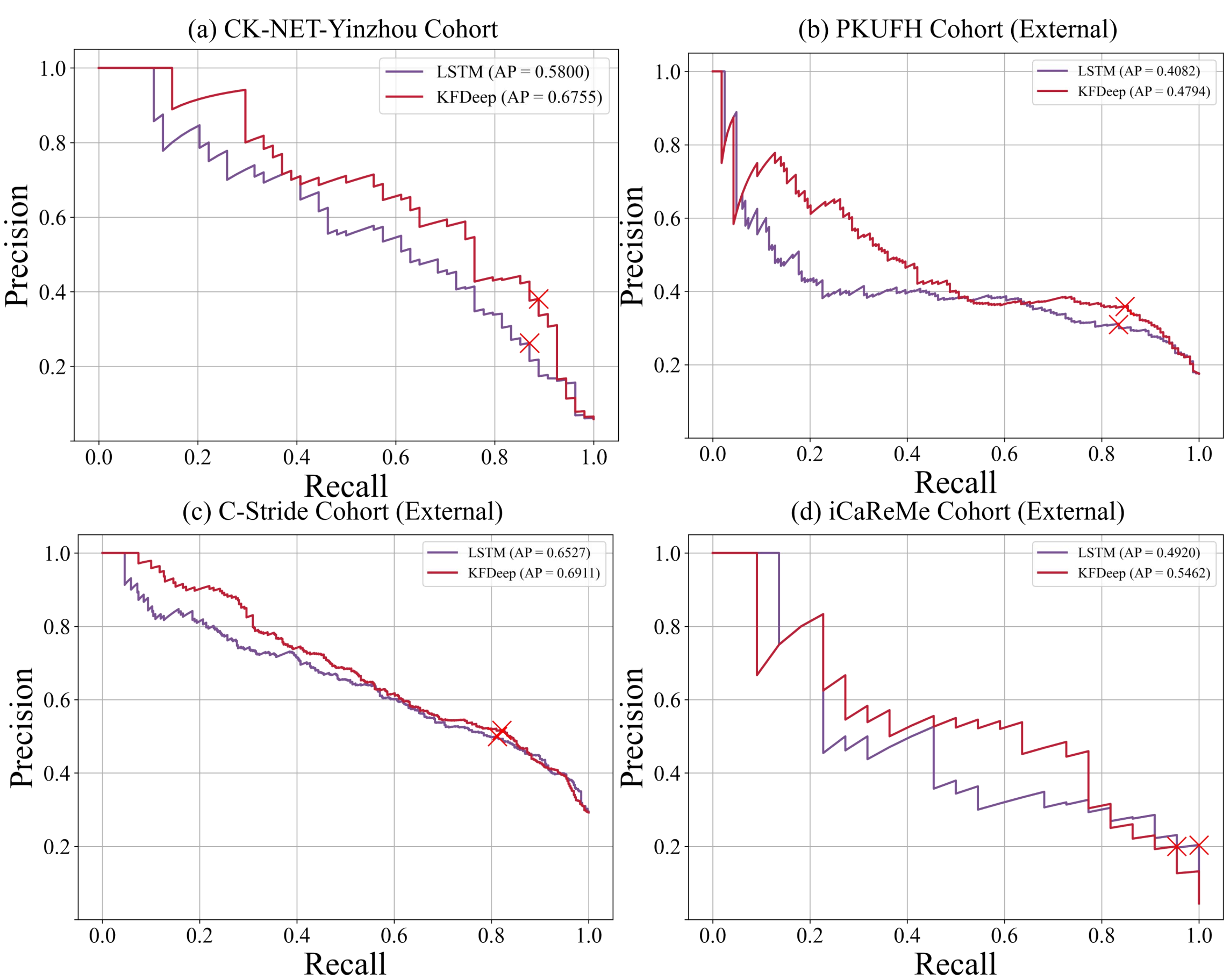

Figure O.1: Precision-recall curves comparing KFDeep and LSTM across four cohorts.

## Appendix P. Statistical Relationship Between Predictors and Model Prediction

We use the model prediction value as the outcome and test the statistical relationship between this value and eight variables. For the six time series variables, the mean of the repeated measurements was used for testing. Table P.1 shows the P values between the model output value and the eight predictors selected in this study. All P values are less than 0.1, and the P values of four indicators are less than 0.05. This indicates that the model output contains information about the predictors.

Table P.1: P Values of Variables and Model Output.

| **Variable** | ***P* value** |
|---|---|
| Blood Calcium | 0.024* |
| HCO3 | <0.001*** |
| Age | <0.001*** |
| Albumin | 0.001** |
| Gender | <0.001*** |
| Blood Phosphorus | <0.001*** |
| eGFR | <0.001*** |
| uACR | <0.001*** |

## Appendix Q. Case Study

We performed a detailed case study to ascertain whether the model's predictions truly reflect the real-time risk of a patient advancing to kidney failure, randomly selecting a patient with more data records than average. By tracking the model's output over successive time intervals, we observed the interplay between these probabilities and changes in the eGFR variable, whose diminishment is indicative of a more advanced stage of CKD. One patient's data, including outputs from the model and eGFR values, is presented in Figure Q.1. We observed that as the eGFR declines and the patient's CKD severity escalates, the model's predicted risk of kidney failure concomitantly increases.

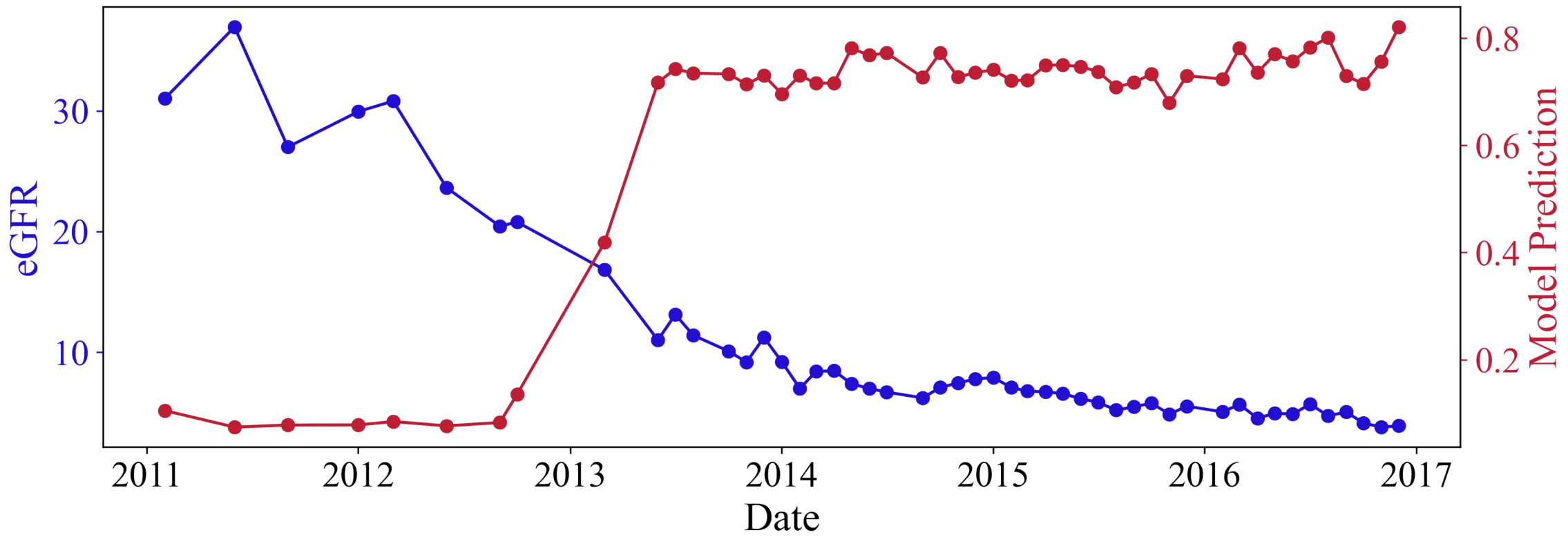

Figure Q.1: Case Study: A Patient's Model Output and eGFR Curve.

## Appendix R. Deployment

To support real-world deployment, we converted the trained KFDeep model parameters into explicit formulas, enabling PyTorch framework-independent deployment. This allows the model to be used without relying on deep learning infrastructure. Building on this, we implemented two deployment settings: (1) we developed a publicly accessible web-based calculator for direct use by clinicians and researchers, and (2) we integrated KFDeep into primary healthcare settings to assist physician decision-making, including deployment across two tertiary hospitals and 15 community health centers in Yinzhou District.

### Appendix R.1. Deployment of the Web-based Calculator

We deploy the KFDeep model on the website https://visdata.bjmu.edu.cn/kfdeep, where users can input a patient's complete historical EHR data to calculate the risk of progression to kidney failure. There are two usage options: one involves entering the data directly on the website, and the other involves uploading the data in a table format. Figure R.1 illustrates how to input data directly, while Figure R.2 demonstrates the use of table-based input. Additionally, we provide a demo dataset, and upon inputting it, users can obtain results as shown in Figure R.3.

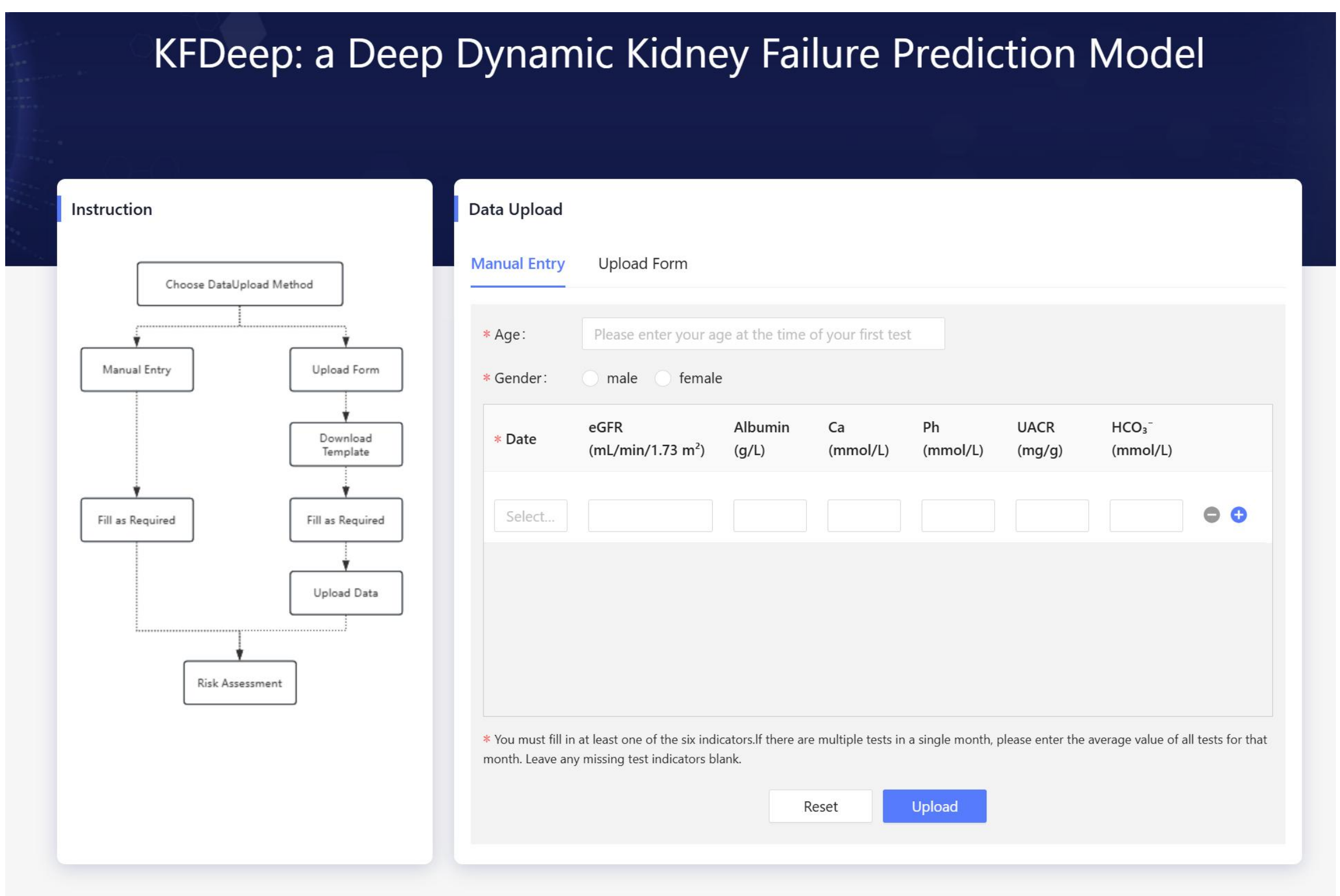

Figure R.1: Website Snapshot: Manual Input. The figure demonstrates how to manually input patient data on the website.

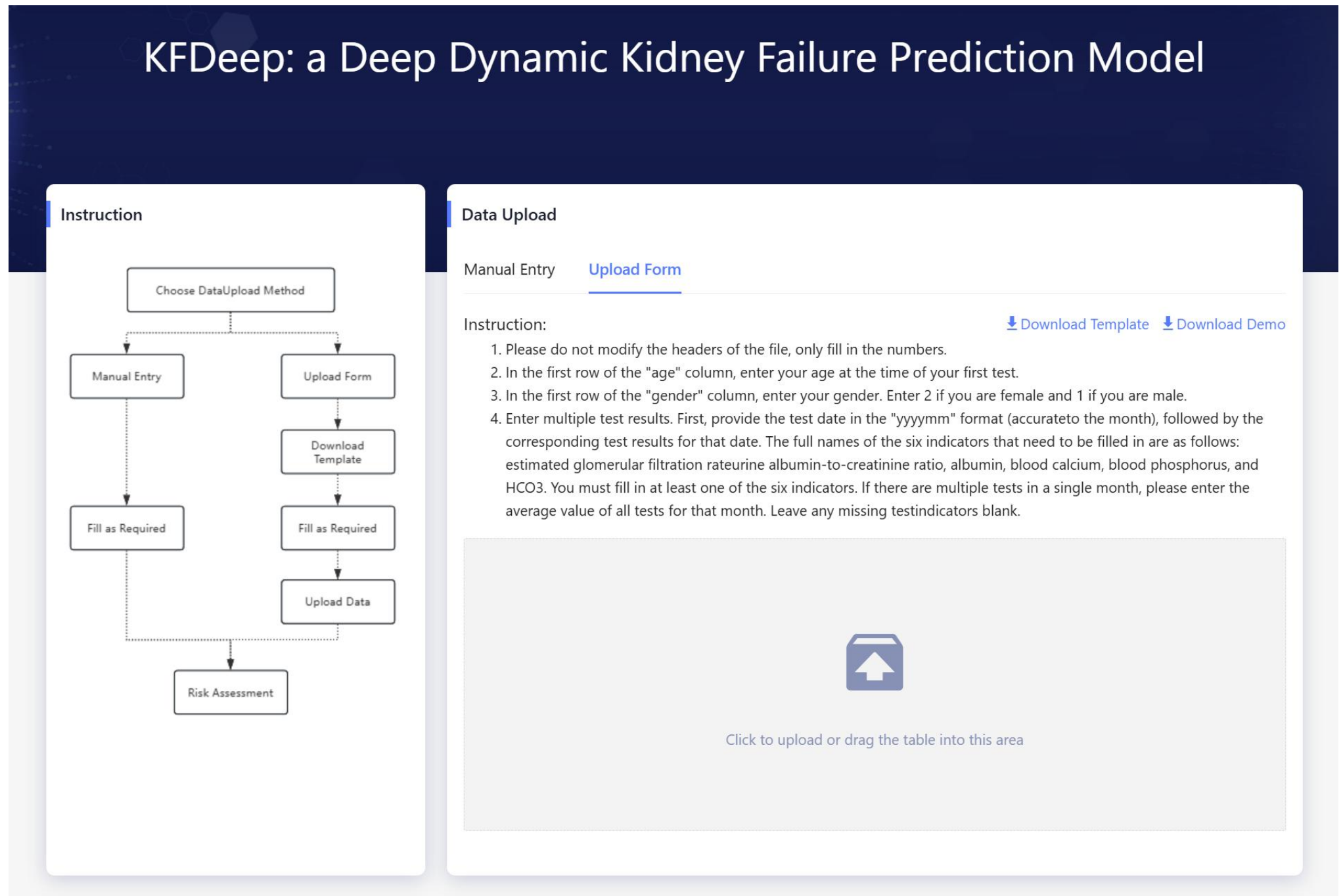

Figure R.2: Website Snapshot: Upload Form. The figure illustrates how to fill out the template table and upload patient data using the table format. A demo file, demo.csv, is also provided on the page for demonstration purposes.

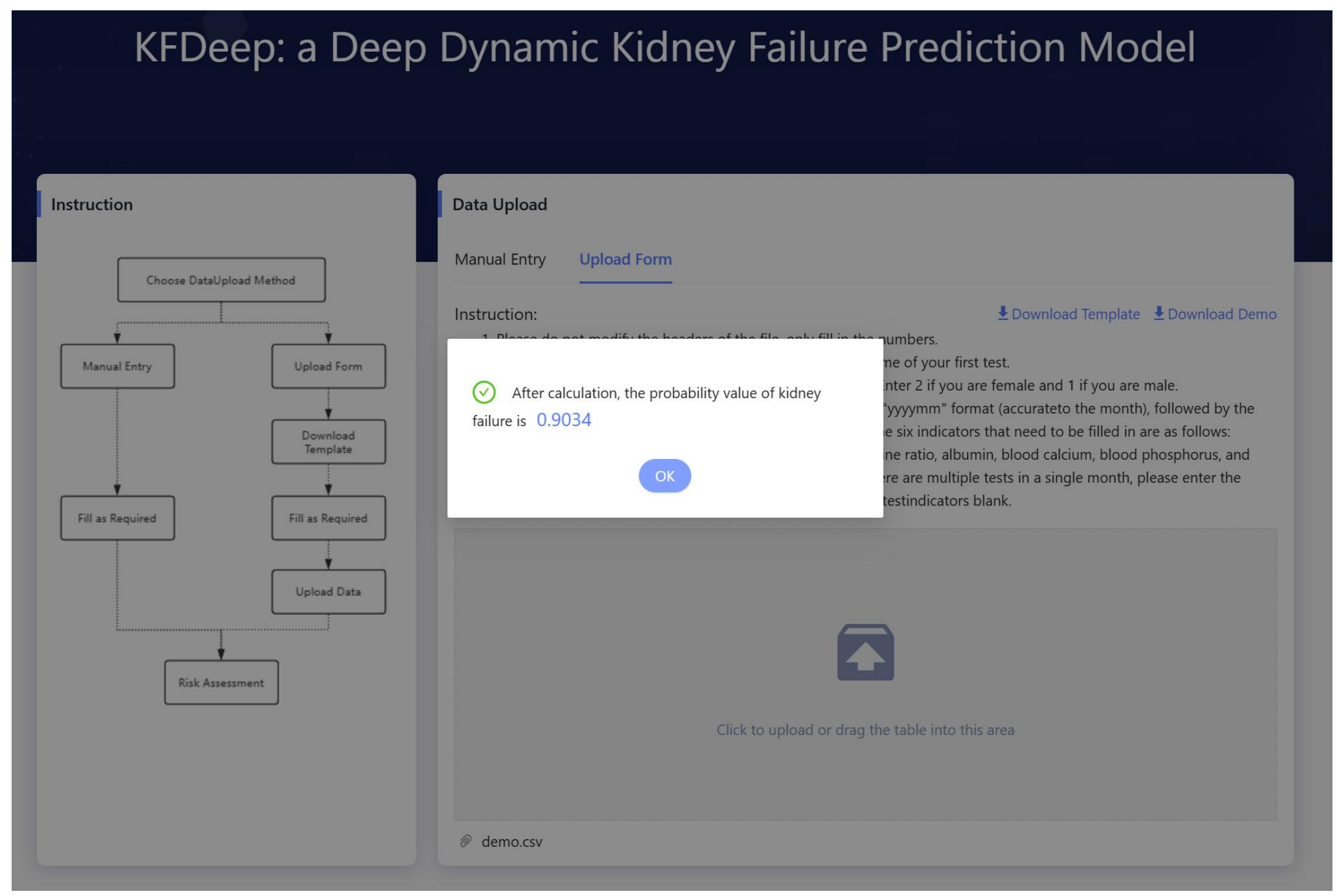

Figure R.3: Website Snapshot: Output. The figure shows the website's output after uploading patient data from the demo.csv file.

## Appendix R.2. Deployment in Primary Healthcare Settings

To extend KFDeep into routine clinical workflows, we collaborated with Yinzhou District to integrate the model into its regional primary healthcare information system. The deployment covered two tertiary hospitals and 15 community health centers, enabling real-time AI-based risk stratification for CKD patients. This pilot integration is informational and non-interventional; it does not change clinical decision making, and it operates under institutional approvals and data-governance procedures.

The system interface (Figure R.4) presents patient demographic details (Name, ID number, Gender, Age), risk level (Low risk, Medium risk, High risk), AI risk stratification (Low risk, Medium risk, High risk), and contract status (Contract status: Signed / Not signed). Risk categories are color-coded to support rapid triage. Clinicians can search by patient ID, name, healthcare institution, and date range, and use functions such as Add, Delete, Record, Risk stratification, and Follow-up.

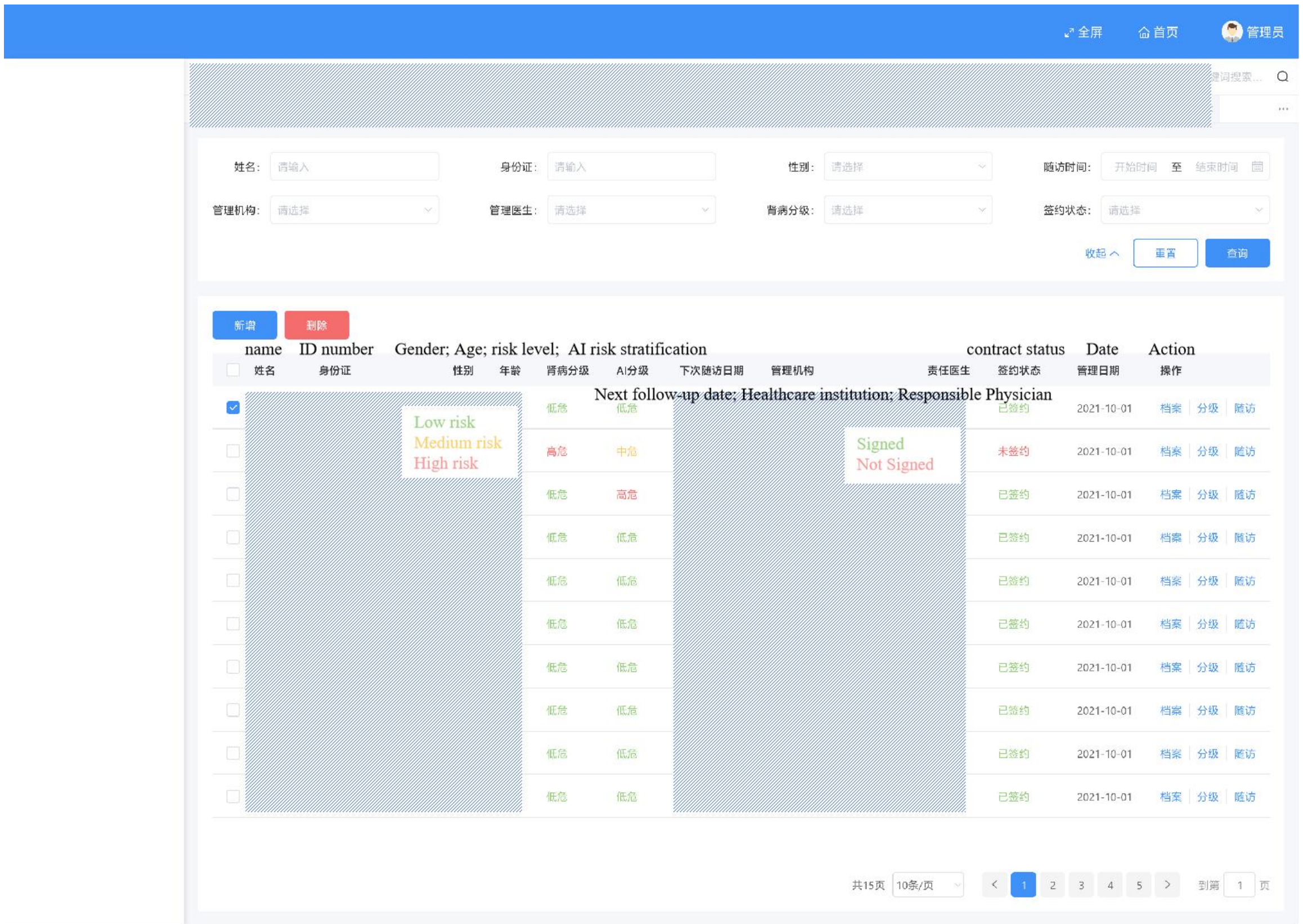

Figure R.4: KFDeep-based CKD risk stratification interface in the primary healthcare information system.

The patient-level view (Figure R.5) supports updating risk levels, scheduling follow-up visits, and documenting management actions (enrolled in management or not currently under management), thereby facilitating continuous and proactive CKD care at the community level.

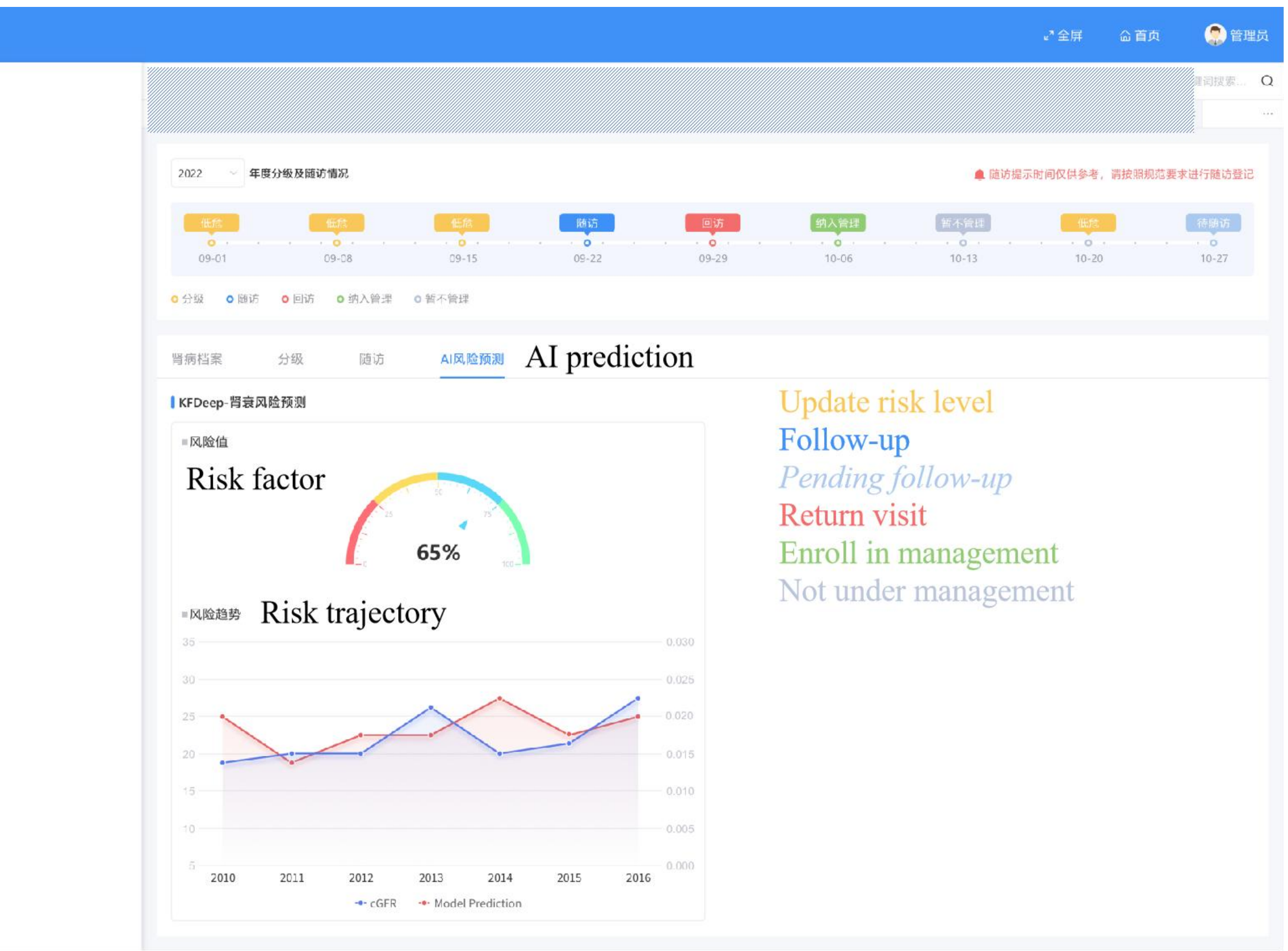

Figure R.5: Patient-level management view in the KFDeep deployment platform.

## Appendix R.3. PyTorch-Free Deployment Implementation

The code for Pytorch-free deployment:

```
import pandas as pd
import numpy as np

def sigmoid(x):
    return 1 / (1 + np.exp(-x))

def tanh(x):
    return np.tanh(x)

def relu(x):
    return np.maximum(0, x)

def calibration_value(value, percentiles=None):
```

```
    if percentiles is None:
        percentiles = [0, 0.001581, 0.003890, 0.004821, 0.006119, 0.007713,

                       0.010107, 0.013142, 0.018956, 0.034004, 1]
    for i in range(len(percentiles) - 1):
        if value <= percentiles[i + 1]:
            mapped_value = (value - percentiles[i]) / percentiles[i + 1] * 0.1 + i * 0.1
            break
    return mapped_value

# model parameters
# The constant parameters W_d, b_d, W_i, b_i, W_f, U_f, b_f, W_g, U_g, b_g, W_o, U_o, b_o, w
eight1, bias1, weight2, and bias2 can all be found on GitHub at the following link: https://github.
com/PKUDigitalHealth/KFDeep.

W_d = [[0.29774544, ..., -0.53426266],
       ...,
       [0.26544067, ..., 3.3814840e-02]]
b_d = [-0.67099166, ..., 0.19365361]
W_i = [[0.4046504, ..., -0.15126696],
       ...,
       [0.02966995, ..., 0.61704564]]
U_i = [[0.257422835, ..., -0.881525099],
       ...,
       [-0.241230682, ..., 0.142124593]]
b_i = [0.13184471, ..., 0.54421222]
W_f = [[0.69906193, ..., -0.77633256],
       ...,
       [-0.19425584, ..., -0.3938809]]

U_f = [[-0.141040936, ..., -0.158538193],
       ...,
       [-9.96147692e-02, ..., 0.444935381]]
b_f = [0.2942414, ..., -0.07398836]
```

```
W_g = [[-0.39057654, ..., 0.03538828],
       ...,
       [-0.4204363, ..., 0.11932335]]
U_g = [[-5.87789714e-02, ..., 0.186236158],
       ...,
       [-0.408906043, ..., -0.461703300]]
b_g = [0, ..., 8.3768040e-02]
W_o = [[-0.1892395, ..., 0.9096555],
       ...,
       [0.05733318, ..., 0.09517419]]
U_o = [[-7.62437433e-02, ..., -0.643919706],
       ...,
       [-0.203064263, ..., 0.375189215]]
b_o = [-0.24270776, ..., 0.29625225]
weight1 = [
    [-0.16387706, ..., -0.06367981],
    ...,
    [0.6001119, ..., -0.14217837]]

bias1 = [-0.18302606, ..., -0.47227636]
weight2 = [[1.1421098, ..., -0.1811245]]
bias2 = [-2.4526906]

hidden_size = 16

if __name__ == '__main__':
    data = pd.read_csv('demo.csv')
    data = data.sort_values(by = 'date', ascending = True)
    data = data.groupby('date').agg(np.mean).reset_index()

    age = data['age'][0]
    gender = data['gender'][0]
    del data['age']
```

```
del data['gender']

interval_list = [0]
for i in range(len(data) - 1):
    interval = int(str(data['date'][i + 1])[2:4]) * 12 + int(str(data['date'][i + 1])[4:6]) - \
            (int(str(data['date'][i])[2:4]) * 12 + int(str(data['date'][i])[4:6]))
    interval_list.append(interval)
data['interval'] = interval_list
del data['date']

Cr = float(110) / 88.4
if gender == 2:
    eGFR = 144 * np.power(Cr / 0.7, -1.209) * np.power(0.993, age)
# for men
else:
    eGFR = 141 * np.power(Cr / 0.9, -1.209) * np.power(0.993, age)
missing_dict = {'egfr': eGFR,
                'albumin': 39,
                'ca': 2,
                'ph': 1,
                'uacr': float(np.exp(2.4738)),
                'hco3': 24.7}
data = data.interpolate()
data = data.bfill()
# if any variable is entirely missing
for idx in missing_dict.keys():
    if pd.isnull(data[idx][0]):
        data[idx] = missing_dict[idx]

epsilon = 1e-6
data['uacr'] = np.log(data['uacr'] + epsilon)

# convert the input type
data = data.values
```

```
# input data
x = data[:, :-1]
delta_t = data[:, -1]

# initialize parameters
h_t = np.zeros(hidden_size)
c_t = np.zeros(hidden_size)

# model
hidden_seq = []
seq_size = len(x)
for t in range(seq_size):
    x_t = x[t, :]
    delta_t_current = delta_t[t] + 0.1
    cs1_tb = np.tanh(c_t @ W_d + b_d)

    # Time decay function
    time_function = 1 / np.power(delta_t_current, 1)

    # T-LSTM Cell computation
    cs2_tb = cs1_tb * time_function
    c_t_b = c_t - cs1_tb
    cx_tb = c_t_b + cs2_tb

    # LSTM Gates computation
    i_t = sigmoid(x_t @ W_i + h_t @ U_i + b_i)
    f_t = sigmoid(x_t @ W_f + h_t @ U_f + b_f)
    g_t = np.tanh(x_t @ W_g + h_t @ U_g + b_g)
    o_t = sigmoid(x_t @ W_o + h_t @ U_o + b_o)

    # Update cell state and hidden state
    c_t = f_t * cx_tb + i_t * g_t
    h_t = o_t * np.tanh(c_t)
```

```
    hidden_seq.append(h_t)

time_output = weight1 @ h_t + bias1
time_output = relu(time_output)
mix_data = np.concatenate((time_output, [age, gender]))
outputs = weight2 @ mix_data + bias2
outputs = sigmoid(outputs)
cal_pred = calibration_value(outputs)
print(f"The risk is {cal_pred}")
```